\documentclass[sigconf]{acmart}

\AtBeginDocument{%
  }

\usepackage{microtype}
\usepackage{graphicx}
\usepackage{subcaption}
\usepackage{booktabs}
\usepackage{adjustbox}
\usepackage{float}
\usepackage{makecell}
\usepackage{multirow}
\usepackage{xcolor}
\usepackage[most]{tcolorbox}
\usepackage{bbding}
\usepackage{pifont}
\usepackage{enumitem}
\usepackage[table]{xcolor}

\usepackage{amsmath}
\usepackage{amssymb}
\usepackage{mathtools}
\usepackage{amsthm}

\theoremstyle{plain}

\theoremstyle{definition}

\theoremstyle{remark}

\definecolor{cvprblue}{rgb}{0.21,0.49,0.74}
\definecolor{myblue}{rgb}{0.9, 0.95, 1}

\newcommand{\ours}{\textsc{AlignGemini}}
\newcommand{\plus}[1]{\textcolor{green!80!black}{\textbf{#1}}}
\newcommand{\minus}[1]{\textcolor{red!80!black}{\textbf{#1}}}

\definecolor{cvprblue}{rgb}{0.21,0.49,0.74}
\definecolor{myblue}{rgb}{0.9, 0.95, 1}

\newcommand{\Strong}{\textcolor{green!60!black}{Strong}}
\newcommand{\Moderate}{\textcolor{orange}{Moderate}}
\newcommand{\Weak}{\textcolor{red}{Weak}}

\newtcolorbox{findingbox}[1]{
  colback=gray!10!white,
  colframe=gray!80!black,
  coltitle=white,
  fonttitle=\bfseries,
  title={#1},
  sharp corners=south,
  boxrule=0.5mm,
  drop shadow
}

\copyrightyear{2026}
\acmYear{2026}
\setcopyright{cc}
\setcctype{by}
\acmConference[MM '26]{Proceedings of the 34th ACM International Conference on Multimedia}{November 10--14, 2026}{Rio de Janeiro, Brazil}
\acmBooktitle{Proceedings of the 34th ACM International Conference on Multimedia (MM '26), November 10--14, 2026, Rio de Janeiro, Brazil}
\acmDOI{10.1145/3767308.3836274}
\acmISBN{979-8-4007-2213-4/2026/11}
\begin{document}

\title[Align, Don't Blend: Orthogonal Supervision Disentanglement As the Key to Generalizable \\ AI-Generated Image Detection]{Align, Don't Blend: Orthogonal Supervision Disentanglement As the Key to Generalizable AI-Generated Image Detection}


\author{Ruoxin Chen}
\email{cusmochen@tencent.com}
\affiliation{
  \institution{Tencent Youtu Lab}
  \city{Shanghai}
  \country{China}
}

\author{Jiahui Gao}
\email{y10250186@mail.ecust.edu.cn}
\affiliation{
  \institution{East China University of Science and Technology}
  \city{Shanghai}
  \country{China}
}

\author{Kaiqing Lin}
\email{linkaiqing2021@email.szu.edu.cn}
\affiliation{
  \institution{Shenzhen University}
  \city{Shenzhen}
  \country{China}
}

\author{Keyue Zhang}
\email{zkyezhang@tencent.com}
\affiliation{
  \institution{Tencent Youtu Lab}
  \city{Shanghai}
  \country{China}
}

\author{Yandan Zhao}
\email{yandanzhao@tencent.com}
\affiliation{
  \institution{Tencent Youtu Lab}
  \city{Shanghai}
  \country{China}
}

\author{Isabel Guan}
\authornote{Project leader.}
\email{eeguan@ust.hk}
\affiliation{
  \institution{The Hong Kong University of Science and Technology}
  \city{Hong Kong}
  \country{China}
}

\author{Taiping Yao}
\email{taipingyao@tencent.com}
\affiliation{
  \institution{Tencent Youtu Lab}
  \city{Shanghai}
  \country{China}
}

\author{Shouhong Ding}
\authornote{Corresponding author.}
\email{ericshding@tencent.com}
\affiliation{
  \institution{Tencent Youtu Lab}
  \city{Shanghai}
  \country{China}
}

\renewcommand{\shortauthors}{Ruoxin Chen et al.}


\begin{abstract}
Recent advances in AI-Generated Image (AIGI) detection have largely focused on model architectures, loss design, and training strategies, while overlooking a more fundamental factor: \emph{how supervision is structured}. Existing detectors typically train on benchmark-defined data that implicitly blends semantic and artifact signals, leaving supervision design unexplored. In this paper, we examine supervision structure through controlled experiments and show that such blended supervision is not merely suboptimal but fundamentally counterproductive. Replacing it with disentangled, orthogonal supervision—paired with inductively matched models—suffices to achieve strong cross-generator generalization. We formalize this as \emph{Orthogonal Supervision Disentanglement} and instantiate it in \textbf{AlignGemini}, a two-branch detector in which 
a VLM receives purely semantic supervision and a vision expert receives purely artifact supervision. With disentangled supervision coupled with inductively matched models, even the simplest training recipe achieves a significant improvement of \textbf{9.5\%} across five in-the-wild benchmarks—trained on images from \textbf{only two generators}, without complex training strategies or diverse generator coverage. We further introduce \textbf{AIGI-Now}, a diagnostic benchmark covering eight contemporary generators, including five closed-source commercial models, with disentangled semantic and pixel evaluation subsets for fine-grained analysis of detector capabilities. Our code is publicly available at \url{https://github.com/GzZning/Align-Don-t-Blend}.
\end{abstract}



\begin{teaserfigure} \vspace{-10pt}
  \centering
  \setkeys{Gin}{width=0.83\textwidth, trim=4 4 4 4, clip}
  \includegraphics{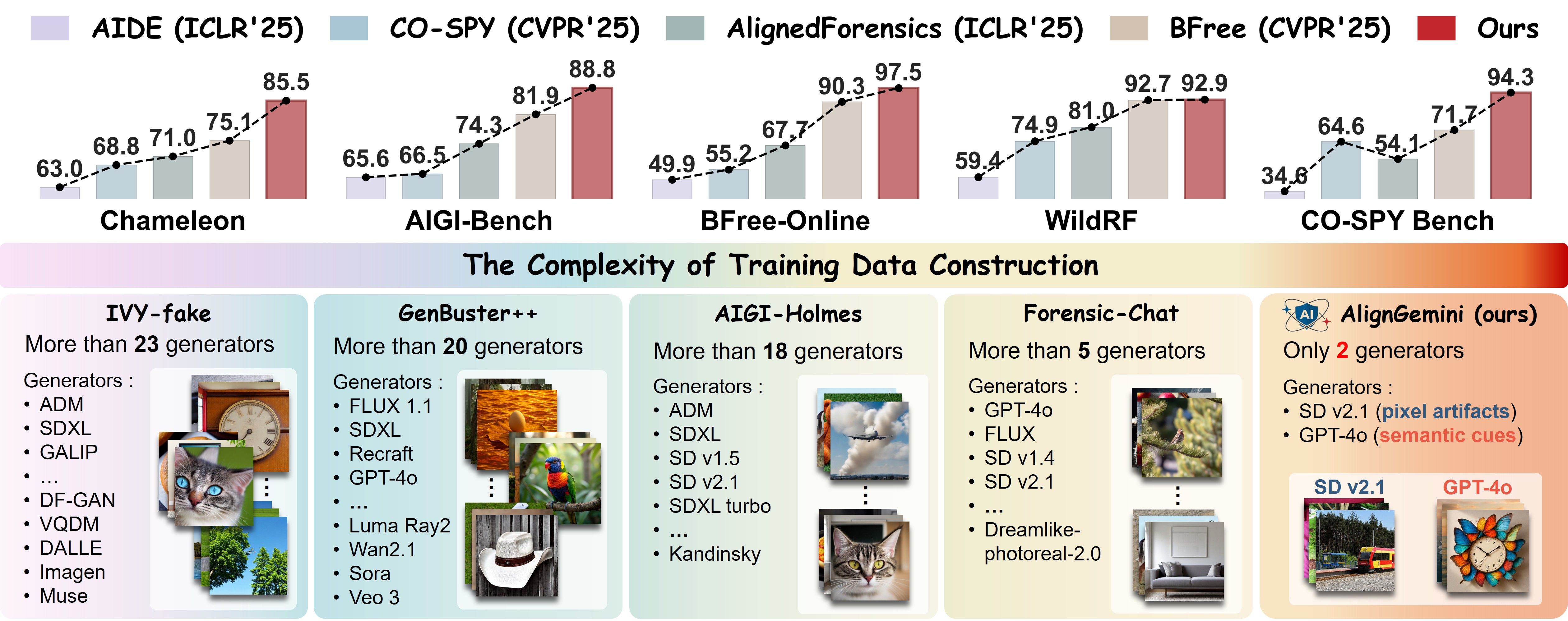} 
    \caption{\textbf{Top:} Performance comparison on five challenging in-the-wild benchmarks. AlignGemini consistently outperforms all existing detectors across all five datasets. \textbf{Bottom:} Comparison of training corpus complexity. AlignGemini relies on a substantially simpler training corpus—only two generators—yet achieves the strongest performance overall. Through orthogonal supervision disentanglement, AlignGemini demonstrates that supervision design, not data scale or generator diversity, is a critical and previously overlooked lever for generalizable AIGI detection.}
  \label{fig:teaser}
\end{teaserfigure}


\maketitle

\section{Introduction}

The rapid progress of generative models~\cite{goodfellow2014generative, ho2020denoising, van2016pixel, yan2025can} has fueled innovation across creative industries, but the growing volume of AI-generated images has heightened misinformation risks, making detection essential. Recent efforts in AIGI detection have primarily concentrated on advancing model architectures and training strategies. Early works~\cite{xu2025fakeshield,chen2024x2,ojha2023towards} explore the potential of pretrained vision foundation models for detection, and subsequent studies enhance detectors via prompt engineering~\cite{ji2025towards}, reinforcement learning~\cite{xu2025avatarshield, zhou2025aigi}, reasoning-oriented mechanisms~\cite{huang2025thinkfake, xia2025mirage, FakeReasoning}, and fusion with expert detectors~\cite{chen2024x2, zhou2025aigi, peng2025mllm, tan2025forenx}. Despite such progress, a more fundamental factor remains largely overlooked: \textbf{supervision design}. Since a detector learns its discriminative capabilities from supervision data, the way supervision is designed is one of the most critical determinants of detection performance.

A line of works~\cite{yan2024sanity,cheng2025co,zhou2025aigi,tan2025forenx,chen2024x2} demonstrate that AIGI detection can be decomposed into two complementary subtasks: high-level semantic consistency checking and low-level pixel-artifact detection. Ignoring either signal leads to systematic blind spots (see Section~\ref{sec:aigi-two-subtasks}). Crucially, these two signals differ fundamentally in nature—semantic cues concern image content, while artifact cues reside in low-level pixel statistics—making it difficult for a single model to handle both effectively. Recent works~\cite{cheng2025co,yan2024sanity,zhou2025aigi} acknowledge this distinction and adjust model architectures accordingly, typically introducing two-branch designs with specialized modules (e.g., CLIP embeddings for semantic features and frequency-domain transforms for artifact cues). \textbf{However, these approaches achieve disentanglement only at the architectural level while leaving supervision blended: standard training corpora do not separate semantic and artifact signals, and both branches are still optimized under a single, conflated objective.}

In this paper, we systematically examine the role of supervision structure through controlled experiments (Section~\ref{paragraph_empirical_analysis}). Our analysis reveals two key findings: (i)~different model families exhibit distinct inductive biases—VLMs are naturally suited to semantic consistency checking, while conventional vision backbones excel at pixel-artifact detection; (ii)~training each model under pure, matched supervision amplifies its inherent inductive strength and yields strong cross-generator generalization, whereas blended supervision degrades performance on the very subtask the model is best equipped to solve. These findings suggest a natural remedy: replacing the prevailing blended supervision paradigm with \emph{disentangled, orthogonal supervision}, in which each signal type is paired with an inductively matched model. We formalize this as \emph{Orthogonal Supervision Disentanglement}, which posits that generalization in AIGI detection is fundamentally governed by the alignment between supervision structure and model inductive bias. Guided by this principle, Section~\ref{sec:aligngemini} introduces \textbf{AlignGemini}, a two-branch detector that instantiates orthogonal supervision with a deliberately simplified training corpus and minimal training overhead, yet achieves a \textbf{9.5\%} improvement across all five in-the-wild benchmarks—demonstrating that supervision design, not data scale, is the key lever for generalizable AIGI detection. 

To support evaluation of disentangled discriminative capabilities beyond aggregate accuracy, we introduce \textbf{AIGI-Now}, a diagnostic benchmark designed for disentangled detector analysis. AIGI-Now spans eight contemporary image generators, and provides two disjoint evaluation subsets per generator to separately assess semantic discrimination and pixel-artifact discrimination. This design facilitates principled development of future AIGI detectors.

\begin{itemize}
    \item \textbf{Orthogonal Supervision Disentanglement.} Through controlled experiments, we show that the prevailing practice of blending semantic and artifact supervision is fundamentally counterproductive, and formalize \emph{Orthogonal Supervision Disentanglement} principle.

    \item \textbf{AlignGemini.} We instantiate this principle in a detector trained on only two generators, with simplified training. AlignGemini achieves a \textbf{9.5\%} improvement across in-the-wild benchmarks, demonstrating that supervision design, not data scale, is the key lever for generalization.

    \item \textbf{AIGI-Now Benchmark.} We introduce a diagnostic benchmark spanning eight contemporary generators (five closed-source) with disentangled semantic and pixel evaluation subsets, enabling fine-grained analysis of detector capabilities beyond aggregate accuracy.
\end{itemize}

\begin{figure*}[t!]
\centering
\includegraphics[width=0.95\textwidth, trim=4 4 4 4, clip]{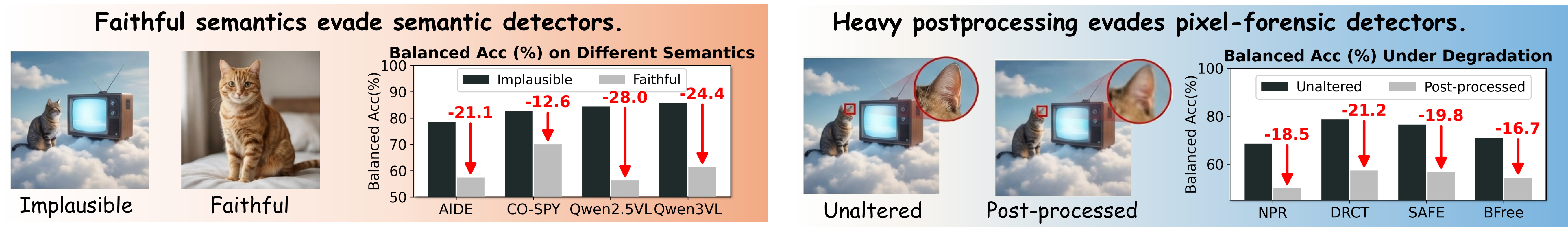}
\vspace{-5pt}
\caption{Illustration of systematic blind spots of semantic and pixel-artifact detectors. \textbf{Left:} semantically-faithful synthetic images evade semantic detectors despite the existence of pixel artifacts. \textbf{Right:} heavy degradations (e.g., compression) destroy the generation pixel trace, reducing artifact detectors' detection rate by 16\%+, even for evidently implausible images.}
\vspace{-5pt}
\label{fig:complementary}
\end{figure*}

\section{Related Work}

\paragraph{Integrated AIGI Detection.}
Early detectors, including CNNSpot~\cite{wang2020cnn}, UnivFD~\cite{ojha2023towards}, and C2P-CLIP~\cite{tan2024c2p}, train a single model on pooled data without disentangling semantic cues from low-level artifacts. Subsequent methods focus more directly on pixel-level artifacts: NPR~\cite{tan2024rethinking} captures upsampling fingerprints through neighboring-pixel relationships, FIRE~\cite{chu2024fire} exploits frequency-guided reconstruction errors, and SAFE~\cite{li2025improving} leverages DCT-based frequency-domain features. Other efforts, including DRCT~\cite{chen2024drct}, BFree~\cite{guillaro2024bias}, and a broad range of related works~\cite{qin2025Scaling,liu2026mirror,liu2025beyond,shen2025dino,liu2026m,cui2026rethinking,lin2025standing,zhou2026breaking,lin2026guard,zhou2026simplicity}, leverage pretrained vision foundation models to improve cross-generator generalization. More recently, VLM-based approaches have shifted their focus toward detecting semantic inconsistencies in AIGI. However, these methods typically require large-scale training corpora and increasingly complex annotation pipelines. For example, FakeShield~\cite{xu2025fakeshield} aggregates six datasets and uses GPT-4o to annotate defective regions and generate corresponding rationales, whereas FakeVLM~\cite{SpotFake} constructs FakeClue from five data sources and employs three VLMs to jointly annotate semantic flaws. IVY-Fake~\cite{zhang2025ivy}, LEGION~\cite{legion}, BusterX++~\cite{wen2025busterx++}, VLForgery~\cite{he2026vlforgery}, and VLM-Defake~\cite{ji2025towards} broaden the coverage of generators and forgery types through expert annotation, cross-LLM validation, or both. Despite their methodological diversity, existing detectors generally emphasize either semantic inconsistencies or low-level artifacts rather than modeling them jointly. 
This separation creates systematic blind spots. \vspace{-5pt}

\paragraph{Disentangled AIGI Detection.}
Recent work recognizes that AIGI detection comprises two complementary subtasks—semantic consistency checking and pixel-artifact detection—and that VLMs are inherently weak at low-level perception~\cite{cheng2025co,jia2024can,yoon2025visual,grover2025huemanity}. To address this, several methods introduce external expert modules alongside a VLM: AIDE~\cite{yan2024sanity} couples the semantic-discriminative CLIP branch with a shallow CNN to inject pixel-level cues; CO-SPY~\cite{cheng2025co} fuses CLIP semantic features with VAE reconstruction residuals to capture artifact signals; X2-DFD~\cite{chen2024x2} augments a large VLM with pixel-forensic expert detectors; AIGI-Holmes~\cite{li2025aigi} pairs a large VLM with CLIP and NPR~\cite{tan2024rethinking}; and AvatarShield~\cite{xu2025avatarshield} concatenates a residual-based expert module. These approaches achieve disentanglement at the \emph{model} level by assigning separate modules to semantic and artifact cues. However, they overlook an equally critical dimension: \emph{supervision}. \textbf{Their training corpora remain blended, entangling signals across branches and limiting specialization.} In contrast, AlignGemini pursues disentanglement at \emph{both} levels jointly: the VLM branch is trained on a purely semantic corpus, while the vision expert is trained exclusively on pixel-artifact cues.

\section{Preliminaries: Disentangling AIGI Detection into Two Subtasks}
\label{sec:aigi-two-subtasks}
AIGI detection decomposes into two complementary subtasks: (i) semantic-consistency verification, which assesses whether image content is semantically plausible, and (ii)~pixel-artifact detection, which identifies low-level traces introduced by the generation process. Each subtask can succeed in isolation but exhibits systematic blind spots, as illustrated in Figure~\ref{fig:complementary}. Semantic detectors can be bypassed by content-faithful synthetic images, while pixel-based detectors are fragile under aggressive post-processing. Each subtask captures a distinct failure mode, and reliance on either alone inevitably leaves systematic vulnerabilities.

\begin{figure*}[t!]
\centering
\includegraphics[width=0.90\textwidth]{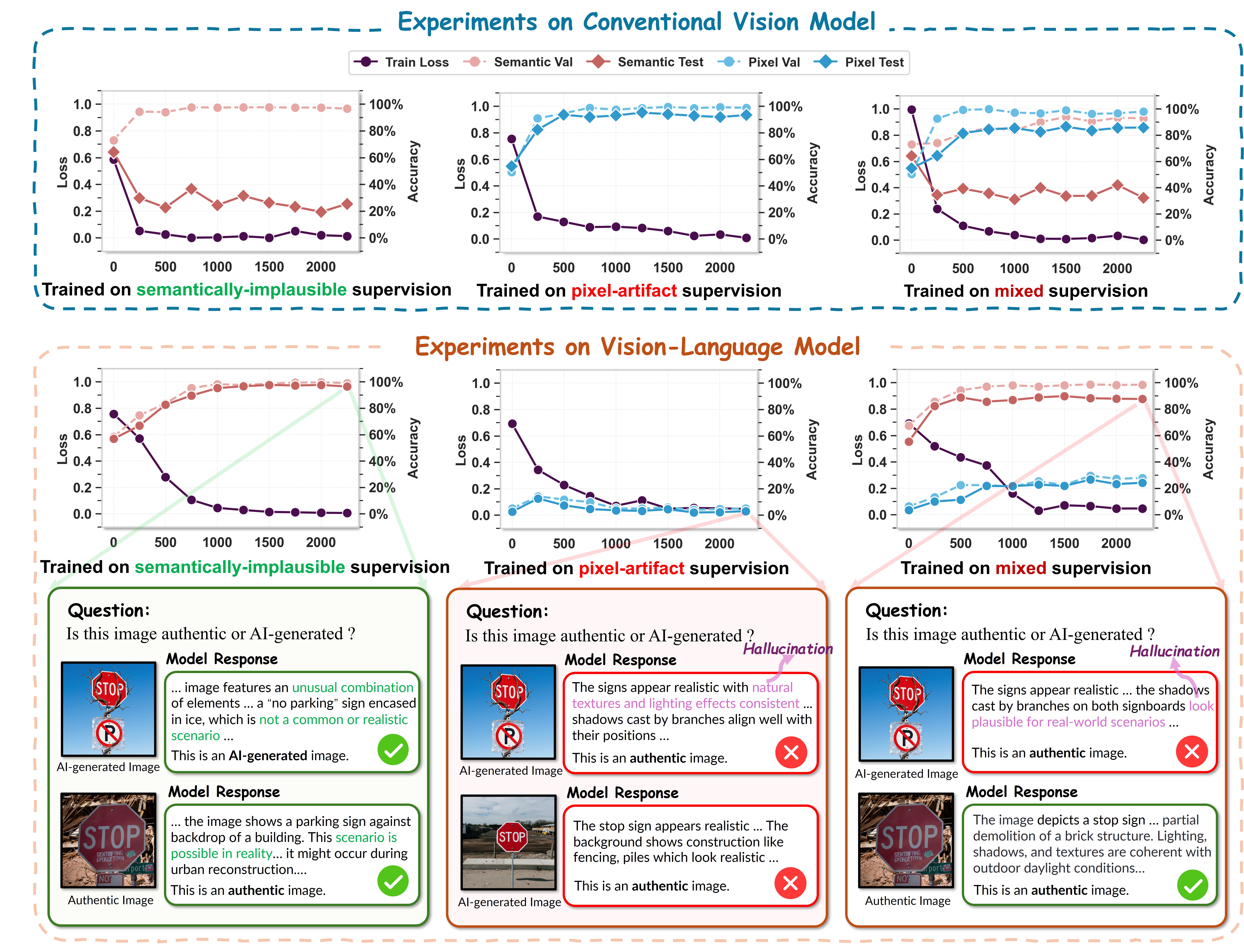}
\vspace{-10pt}
\caption{A controlled study on the effect of supervision structure on models with different inductive strengths. "Semantic/Pixel Val" denotes images generated by the same generator used in training. "Semantic/Pixel Test" denotes images from unseen generators. \textbf{Top:} DINOv2 trained with semantic supervision generalizes poorly to unseen semantic inconsistencies, whereas pixel-artifact supervision yields strong generalization on pixel-artifact tests. \textbf{Bottom:} Qwen-VL trained with semantic supervision robustly detects semantic flaws from unseen generators, while pixel-artifact supervision fails to improve its sensitivity to low-level artifacts. These results reveal a clear inductive preference of each model toward different signal types. \textbf{Rightmost column:} Blended semantic–pixel supervision degrades each model's inherent strengths. \textbf{The results indicate that semantic and pixel subtasks align with different inductive biases. Joint supervision cannot reconcile this mismatch, whereas task–model alignment yields superior generalization.}} \vspace{-10pt}
\label{fig:misalignment}
\end{figure*}

\section{Empirical Foundation: How Supervision Disentanglement Affects Detection}
\label{paragraph_empirical_analysis}
Having established that AIGI detection decomposes into two subtasks, we now examine the impact of supervision design on models with distinct inductive biases. We construct three training sets: (i)~\emph{Pixel-artifact set}, where synthetic images are generated via VAE reconstruction to preserve semantics while altering pixel statistics; (ii)~\emph{Semantically-implausible set}, where images contain semantic inconsistencies and are aggressively post-processed to suppress low-level cues; and (iii)~a \emph{Blended set} combining both. We train a VLM (Qwen-VL) and a conventional vision model (DINOv2) under each supervision regime and observe two key findings (Figure~\ref{fig:misalignment}). First, we observe that \textbf{different models have different inductive strengths}:  VLMs consistently excel at semantic-flaw detection, and semantic supervision further strengthens this capability with strong cross-generator generalization; however, they remain largely insensitive to pixel-level artifacts, even under explicit pixel supervision. In contrast, DINOv2 effectively captures pixel artifacts and generalizes well across generators, but performs poorly on semantic inconsistency detection. Second, \textbf{keeping supervision disentangled is critical}: pure, matched supervision amplifies each model's inherent specialization, whereas blending both signal types into a single model does not preserve but rather severely degrades each model's strengths.

\section{Methodology}
\label{sec:aligngemini}
\begin{figure*}[t!]
\centering
\includegraphics[width=.9\textwidth]{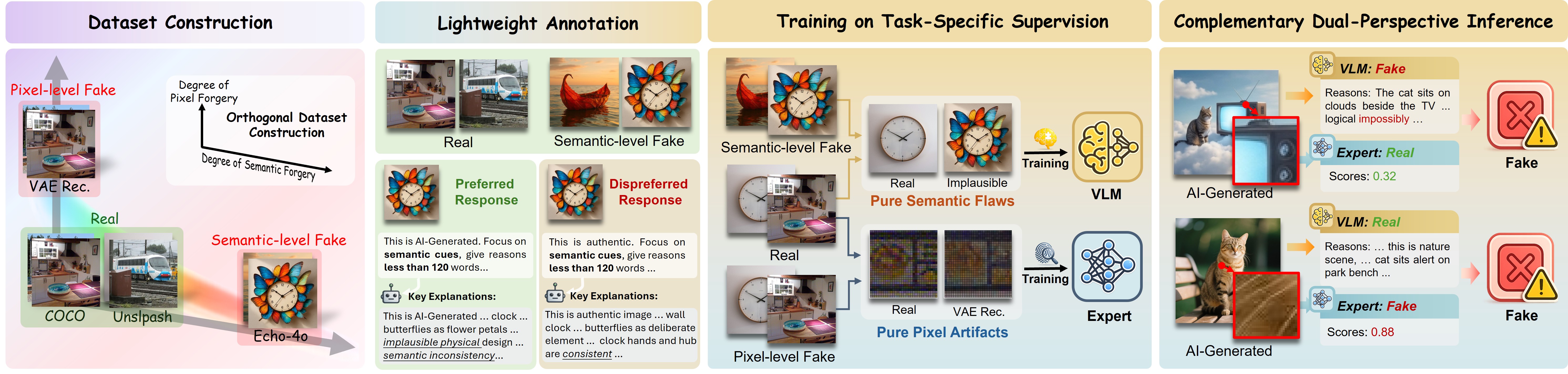} \vspace{-5pt}
\caption{Illustration of the pipeline. \textbf{AlignGemini adopts a deliberately simple yet effective design} by constructing two orthogonal supervision sets. For semantic supervision, a pretrained VLM generates DPO-style labels explicitly constrained to semantic cues. The VLM and the expert are then fine-tuned separately, aligning each model with its corresponding subtask to encourage complementary specialization. At inference, AlignGemini jointly evaluates semantic and pixel-level cues.} \vspace{-10pt}
\label{fig:pipeline}
\end{figure*}

Based on the findings from the controlled study above, we establish two key principles: (1) supervision must be matched to a model's inductive bias, and (2) purely disentangled supervision maximizes each branch's specialization. Guided by these principles, we propose AlignGemini (Figure~\ref{fig:pipeline}), a detector trained with only a simplified training corpus and lightweight fine-tuning. \textbf{The key distinction from prior work is that we deliberately construct disentangled, orthogonal supervision data from scratch.} We do not pursue large-scale training data, diverse generator coverage, or expert annotations; instead, simple yet disentangled, orthogonal supervision suffices to achieve remarkable in-the-wild performance.


\paragraph{Orthogonal Supervision Disentanglement.}
Orthogonal supervision disentanglement emphasizes purity and orthogonality. We construct two supervision sets in which semantic and pixel-artifact cues are disjoint by design.
\textbf{(i) Semantic supervision set.} The goal of semantic supervision is to ensure that semantically implausible content is present for the model to learn from, while eliminating low-level cues to preserve purity. To this end, we collect 5,000 semantically implausible synthetic images from Echo-4o~\cite{ye2025echo}, whose generation prompts have been injected with semantic flaw keywords to guarantee the presence of semantic inconsistencies. These are coupled with 5,000 real images from Unsplash~\cite{unsplash_data_2025} to train the VLM branch. To prevent leakage of low-level cues, both real and synthetic images undergo online randomized heavy post-processing during training, including resolution downsampling, JPEG compression, noise injection, and upsampling to the original size. This process perturbs pixel statistics; in particular, the downsampling–upsampling operation disrupts neighborhood pixel relationships. For annotation, we adopt a Direct Preference Optimization (DPO) protocol by prepending a fixed verdict prefix (\emph{"This is an AI-generated image."} / \emph{"This is an authentic image."}) and eliciting paired responses from Qwen2.5-VL-72B~\cite{qwen25} to form preferred–dispreferred samples.
\textbf{(ii) Pixel-artifact supervision set.} This set consists of natural images paired with their Stable Diffusion 2.1 VAE reconstructions, forming semantically matched pairs, following the construction pipeline of DDA~\cite{chen2025dual}. This design removes semantic bias and isolates low-level discrepancies for pixel-artifact supervision.
By keeping the two supervision sets orthogonal, AlignGemini allows each branch to specialize in its respective subtask, thereby improving the effectiveness of the integrated detector. Section~\ref{sec:experiment} verifies that such orthogonality is critical for achieving robust generalization.

\begin{figure*}[t!]
\centering
\includegraphics[width=.8\textwidth, trim=10 10 10 10, clip]{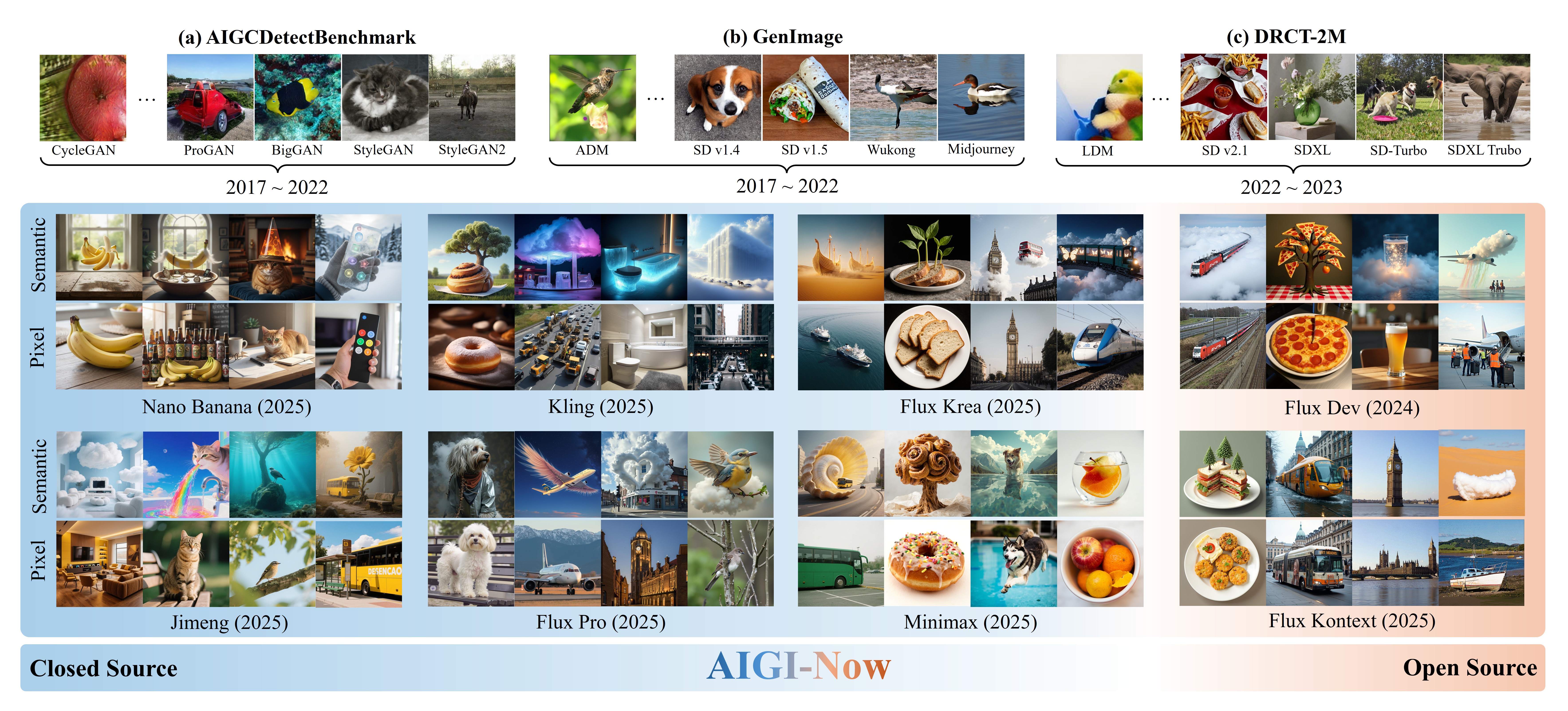}
\vspace{-5pt}
\caption{\textbf{Comparison between AIGI-Now and existing benchmarks.} \textbf{Top:} Existing benchmarks mainly rely on pre-2023 open-source generators. By contrast, real-world usage is dominated by newer commercial, closed-source models, creating a clear gap between the existing benchmark data and real-world fakes. \textbf{Bottom:} Our proposed AIGI-Now is constructed from the latest generators, including five commercial, closed-source models with unknown architectures, enabling rigorous evaluation of detectors’ cross-architecture generalization.}
\label{fig:aigi-now} \vspace{-5pt}
\end{figure*}

\vspace{-5pt}

\paragraph{Training of AlignGemini.}
AlignGemini enforces task–model alignment by pairing each subtask with a model suited to its inductive strengths: a VLM for semantic discrimination and a vision expert for pixel-forensic detection. The VLM branch is fine-tuned via DPO on the semantically implausible supervision, while the expert branch is fine-tuned on the pixel-artifact supervision. 

\vspace{-5pt}

\paragraph{Inference of AlignGemini.}
At inference time, AlignGemini evaluates an input image using two complementary signals: semantic consistency and pixel-level artifact evidence. The underlying decision rule is conservative: since authentic images are expected to be benign along both dimensions, the presence of either semantic inconsistencies or pixel-level artifacts is sufficient to indicate synthesis. Accordingly, AlignGemini adopts a logical OR fusion, classifying an image as AI-generated if either the semantic or artifact branch produces a positive prediction.
For semantically interpretable cues, the VLM branch provides linguistically grounded reasoning. Pixel-level artifacts, which are not expressible in natural language, are assessed via the expert’s output logits.

\vspace{-5pt}
\section{Proposed Benchmark: AIGI-Now}

\paragraph{AIGI-Now: A Diagnostic Benchmark for Disentangled Evaluation.}
AIGI-Now (Figure~\ref{fig:aigi-now}) covers eight representative generators: Nano Banana, Jimeng, Kling, Minimax, Flux Pro, Flux Krea, Flux Kontext, and Flux Dev. For each generator, we sample 1,000 real images from the COCO test set~\cite{mscoco} and use their captions—along with deliberately perturbed variants—to synthesize images that are either semantically faithful or semantically implausible. This yields two complementary diagnostic subsets that explicitly disentangle semantic reasoning from low-level forensic evidence. The \emph{semantic-discriminative} subset pairs real images with semantically implausible synthetic images, all aggressively post-processed to suppress pixel-level cues, thereby isolating semantic signals. The \emph{pixel-artifact-discriminative} subset pairs real images with semantically faithful synthetic images, eliminating semantic shortcuts and forcing reliance on pixel-level artifacts.

\begin{table*}[t!]
\centering
\caption{Overall comparison of balanced accuracy on five in-the-wild benchmarks.} 
\label{tab:compare-in-the-wild}
\vspace{-10pt}
\setlength{\tabcolsep}{3.5pt}
\renewcommand{\arraystretch}{1.05}
\newcommand{\venue}[1]{\textsubscript{\textcolor{blue}{(#1)}}}
\newcommand{\na}{\makecell[c]{--}}

\begin{adjustbox}{width=0.95\linewidth}
\begin{tabular}{l c cccc ccc cccccc cc}
\toprule
\multirow{2}{*}{Method} & \multirow{2}{*}{Chameleon} 
& \multicolumn{4}{c}{WildRF} 
& \multicolumn{3}{c}{AIGI-Bench} 
& \multicolumn{6}{c}{CO-SPY-Bench/in-the-wild} 
& \multirow{2}{*}{BFree-Online}
& \multirow{2}{*}{Avg.} \\
\cmidrule(lr){3-6} \cmidrule(lr){7-9} \cmidrule(lr){10-15}
& & Facebook & Reddit & Twitter & Avg.
& CommunityUI & SocialRF & Avg.
& Civitai & DALL·E 3 & instavibe.ai & Lexica & Midjourney-v6 & Avg. \\
\midrule

NPR\venue{CVPR'24}\cite{tan2024rethinking}
& 59.9 & 78.1 & 61.0 & 51.3 & 63.5 & 54.2 & 58.6 & 56.4 & 0.2 & 4.5 & 0.4 & 4.6 & 82.5 & 18.4 & 40.4 & 47.7 $\pm$ 16.6 \\

UnivFD\venue{CVPR'23}\cite{ojha2023towards}
& 50.7 & 49.1 & 60.2 & 56.5 & 55.3 & 51.4 & 54.6 & 53.0 & 3.0 & 10.3 & 0.3 & 1.2 & 20.8 & 7.1 & 48.8 & 43.0 $\pm$ 18.1 \\

FatFormer\venue{CVPR'24}\cite{liu2024forgery}
& 51.2 & 54.1 & 68.1 & 54.4 & 58.9 & 51.9 & 55.4 & 53.7 & 0.0 & 1.7 & 0.0 & 2.8 & 32.2 & 7.3 & 50.0 & 44.2 $\pm$ 18.7 \\

SAFE\venue{KDD'25}\cite{li2024improving}
& 59.9 & 50.9 & 74.1 & 37.5 & 54.2 & 54.5 & 58.4 & 56.4 & 0.3 & 0.6 & 0.2 & 0.0 & 93.4 & 18.9 & 50.5 & 48.0 $\pm$ 14.9 \\

C2P-CLIP\venue{AAAI'25}\cite{tan2024c2p} 
& 51.1 & 54.4 & 68.4 & 55.9 & 59.6 & 51.0 & 64.9 & 57.9 & 2.5 & 21.9 & 0.3 & 7.6 & 17.0 & 9.9 & 50.0 & 45.7 $\pm$ 18.3 \\

AIDE\venue{ICLR'25}\cite{yan2024sanity}
& 63.0 & 57.8 & 71.5 & 48.8 & 59.4 & 66.2 & 64.9 & 65.6 & 15.1 & 33.1 & 6.9 & 26.6 & 91.3 & 34.6 & 49.9 & 54.5 $\pm$ 11.3 \\

DRCT\venue{ICML'24}\cite{chen2024drct}
& 56.6 & 58.3 & 56.4 & 50.5 & 55.1 & 54.8 & 54.0 & 54.4 & 51.9 & 90.8 & \textbf{85.3} & \textbf{99.6} & 76.1 & \underline{80.7} & 56.1 & 60.6 $\pm$ 10.1 \\

AlignedForensics\venue{ICLR'25}\cite{rajan2025aligned}
& 71.0 & 91.6 & 70.9 & 80.6 & 81.0 & 75.6 & 73.0 & 74.3 & 97.8 & 68.0 & 8.0 & 0.5 & 96.4 & 54.1 & 67.7 & 69.6 $\pm$ 8.9\\

CO-SPY\venue{CVPR'25}\cite{cheng2025co}
& 68.8 & 72.2 & 77.7 & 74.8 & 74.9 & 65.2 & 67.7 & 66.5 & 96.3 & 69.0 & 37.6 & 68.8 & 51.4 & 64.6 & 55.2 & 66.0 $\pm$ 6.4\\

Forensic-Chat (Qwen2.5-VL)\cite{lin2025seeing} & - & 77.8 & 83.4 & 82.1 & 81.1 & 74.6 & \textbf{89.8} & \underline{82.2} & - & - & - & - & - & - & - & - \\

BFree \textsubscript{\textcolor{blue}{(CVPR'25)}} \cite{guillaro2024bias} 
& \underline{75.1} & \textbf{94.7} & \underline{85.5} & \textbf{97.9} & \underline{92.7} & \underline{79.1} & {84.8} & {81.9} & \underline{98.7} & \underline{96.4} & 14.3 & 50.5 & \underline{98.4} & 71.7 & \underline{90.3} & \underline{82.3} $\pm$ 8.2\\

\midrule
\rowcolor{myblue}

\rowcolor{myblue}
\textbf{AlignGemini} 
& \textbf{85.5} 
& \underline{92.2} & \textbf{90.9} & \underline{95.5} & \textbf{92.9} 
& \textbf{91.9} & \underline{85.6} & \textbf{88.8} 
& \textbf{100.0} & \textbf{99.6} & \underline{75.0} & \underline{96.8} & \textbf{99.9} & \textbf{94.3} 
& \textbf{97.5} 
& \textbf{91.8} $\pm$ 4.2 (\plus{+9.5})
\\ 

\bottomrule
\end{tabular}
\end{adjustbox}
\vspace{-5pt}
\end{table*}

\begin{table*}[tb!]
\centering
\caption{Overall comparison of balanced accuracy on proposed AIGI-Now datasets.}
\vspace{-10pt}
\label{tab:compare-aiginow}

\begin{adjustbox}{width=0.95\linewidth}
\newcommand{\sem}{\textit{Sem}}
\newcommand{\pix}{\textit{Pix}}

\begin{tabular}{l*{18}{c}}
\toprule

& \multicolumn{2}{c}{Nano Banana}
& \multicolumn{2}{c}{Jimeng}
& \multicolumn{2}{c}{Kling}
& \multicolumn{2}{c}{Minimax}
& \multicolumn{2}{c}{Flux Pro}
& \multicolumn{2}{c}{Flux Krea}
& \multicolumn{2}{c}{Flux Dev}
& \multicolumn{2}{c}{Flux Kontext}
& \multicolumn{2}{c}{Avg.} \\

\cmidrule(lr){2-3}
\cmidrule(lr){4-5}
\cmidrule(lr){6-7}
\cmidrule(lr){8-9}
\cmidrule(lr){10-11}
\cmidrule(lr){12-13}
\cmidrule(lr){14-15}
\cmidrule(lr){16-17}
\cmidrule(lr){18-19}

Method
& \pix & \sem
& \pix & \sem
& \pix & \sem
& \pix & \sem
& \pix & \sem
& \pix & \sem
& \pix & \sem
& \pix & \sem
& \pix & \sem \\

\midrule

NPR \textsubscript{\textcolor{blue}{(CVPR'24)}}
\cite{tan2024rethinking}
& 84.4 & 49.9
& 41.9 & 49.9
& 88.3 & 49.9
& 52.1 & 50.0
& 42.7 & 50.0
& 52.3 & 50.0
& 89.3 & 50.0
& 79.2 & 49.9
& 66.3 $\pm$ 19.5 & 50.0 $\pm$ 0.1 \\

UnivFD \textsubscript{\textcolor{blue}{(CVPR'23)}}
\cite{ojha2023towards}
& 51.0 & 51.4
& 50.7 & 52.3
& 51.4 & 53.9
& 50.5 & 52.1
& 51.9 & 52.0
& 51.0 & 52.2
& 63.6 & 49.9
& 57.5 & 55.9
& 53.5 $\pm$ 4.4 & 52.5 $\pm$ 1.7 \\

FatFormer \textsubscript{\textcolor{blue}{(CVPR'24)}}
\cite{liu2024forgery}
& 53.4 & 49.9
& 48.4 & 49.9
& 54.8 & 50.0
& 49.1 & 50.0
& 48.4 & 50.0
& 49.5 & 50.0
& 52.4 & 50.0
& 63.6 & 49.9
& 52.5 $\pm$ 4.8 & 50.0 $\pm$ 0.0 \\

SAFE \textsubscript{\textcolor{blue}{(KDD'25)}}
\cite{li2024improving}
& \textbf{99.3} & 50.1
& 50.0 & 50.5
& \textbf{99.6} & 50.0
& 49.8 & 50.1
& 49.9 & 50.5
& 51.2 & 50.1
& \textbf{99.8} & \underline{75.3}
& \textbf{98.8} & \underline{83.4}
& 74.8 $\pm$ 24.6 & 57.5 $\pm$ 12.8 \\

C2P-CLIP \textsubscript{\textcolor{blue}{(AAAI'25)}}
\cite{tan2024c2p}
& 50.0 & 50.0
& 49.8 & 50.0
& 50.7 & 49.9
& 50.3 & 50.0
& 52.0 & 50.0
& 50.1 & 50.0
& 52.6 & 50.0
& 60.7 & 50.0
& 52.0 $\pm$ 3.4 & 50.0 $\pm$ 0.0 \\

AIDE \textsubscript{\textcolor{blue}{(ICLR'25)}}
\cite{yan2024sanity}
& \underline{98.9} & 51.8
& 63.9 & 51.4
& \underline{98.2} & 55.4
& 51.4 & 54.1
& 60.1 & 53.8
& 50.4 & 56.9
& \underline{99.1} & 59.0
& \underline{97.9} & 80.6
& 77.5 $\pm$ 21.4 & 57.9 $\pm$ 8.9 \\

DRCT \textsubscript{\textcolor{blue}{(ICML'24)}}
\cite{chen2024drct}
& 68.9 & 57.7
& \underline{89.3} & 57.7
& 66.9 & 56.8
& 78.4 & 57.1
& \underline{83.0} & 58.2
& \textbf{91.6} & 58.2
& 86.5 & 58.6
& 86.9 & 55.5
& \underline{81.4} $\pm$ 8.7
& 57.5 $\pm$ 0.9 \\

AlignedForensics
\textsubscript{\textcolor{blue}{(ICLR'25)}}
\cite{rajan2025aligned}
& 88.0 & 50.3
& 82.1 & 50.0
& 69.6 & 50.2
& 56.7 & 50.3
& 65.5 & 49.9
& 59.5 & 50.1
& 78.6 & 50.2
& 65.0 & 50.0
& 70.6 $\pm$ 10.5
& 50.1 $\pm$ 0.1 \\

CO-SPY \textsubscript{\textcolor{blue}{(CVPR'25)}}
\cite{cheng2025co}
& 74.1 & \underline{58.3}
& 83.0 & \underline{76.2}
& 88.3 & \underline{78.8}
& \underline{77.9} & \underline{65.6}
& 77.3 & \underline{72.7}
& \underline{89.1} & \underline{74.0}
& 88.5 & 74.9
& 65.1 & 66.0
& 80.4 $\pm$ 7.9
& \underline{70.8} $\pm$ 6.4 \\

BFree \textsubscript{\textcolor{blue}{(CVPR'25)}}
\cite{guillaro2024bias}
& 69.6 & 52.8
& 75.6 & 54.5
& 86.1 & 60.7
& 56.2 & 49.9
& 71.8 & 54.5
& 59.5 & 51.5
& 74.8 & 54.5
& 77.1 & 53.4
& 71.3 $\pm$ 9.0
& 54.0 $\pm$ 3.0 \\

\midrule

\rowcolor{myblue}
\textbf{AlignGemini}
& 96.0 & \textbf{95.6}
& \textbf{96.2} & \textbf{95.6}
& 97.3 & \textbf{94.9}
& \textbf{91.3} & \textbf{89.8}
& \textbf{89.9} & \textbf{93.2}
& 83.0 & \textbf{92.6}
& 96.6 & \textbf{95.3}
& 86.6 & \textbf{88.0}
& \textbf{92.1} $\pm$ 5.0 (\plus{+10.7})
& \textbf{93.1} $\pm$ 2.7 (\plus{+22.3}) \\

\bottomrule
\end{tabular}
\end{adjustbox}
\end{table*}

\vspace{-5pt}
\section{Experiments}
\label{sec:experiment}

\paragraph{Experimental setup.}
We evaluate all methods on nine benchmarks, organized into three suites: 
(i) \textbf{five in-the-wild benchmarks}—Chameleon~\cite{yan2024sanity}, WildRF~\cite{cavia2024real}, AIGI-Bench~\cite{li2025artificial}, Co\mbox{-}SPY\mbox{-}Bench (in-the-wild)~\cite{cheng2025co}, and BFree\mbox{-}Online~\cite{guillaro2024bias}; 
(ii) \textbf{AIGI-Now}; and 
(iii) \textbf{three self-synthesized benchmarks}—GenImage~\cite{zhu2023genimage}, DRCT\mbox{-}2M~\cite{chen2024drct}, and AIGCDetectBenchmark~\cite{zhong2024patchcraft}. Additional experimental results and ablation studies are provided in the Appendix. \vspace{-10pt}

\paragraph{Implementation Details.}
\emph{(i) VLM:} Qwen2.5\text{-}VL\text{-}7B~\cite{qwen25} is used as the VLM backbone and fine\text{-}tuned on a \emph{pure semantic} set of 5{,}000 randomly sampled Unsplash~\cite{unsplash_data_2025} real images and 5{,}000 randomly sampled Echo-4o~\cite{ye2025echo} synthetic images. 
During DPO fine-tuning, we apply LoRA (rank ${=}16$, $\alpha {=} 32$), using $lr =1{\times}10^{-6}$, batch size $8$, and DPO coefficient $\beta {=} 0.05$ for one epoch. 
\emph{(ii) Expert branch:} we adopt a DINOv2 with a classification head, fine-tuned using LoRA (rank ${=}8$, $\alpha {=}1.0$) with learning rate $1{\times}10^{-4}$ and batch size $16$ on a pure pixel-artifact corpus constructed from VAE reconstructions: 11{,}8 k COCO training data paired with VAE-reconstructed data. At inference, an image is classified as real only if both the VLM and the expert predict real; otherwise, it is synthetic. 

\begin{table*}[t]
\centering

\begin{minipage}[t]{0.48\textwidth}
\centering
\caption{Comparison on three self-synthesized datasets.}
\vspace{-8pt}
\label{tab:compare-self-synthesized}
\begin{adjustbox}{width=\linewidth}
\begin{tabular}{lcccc}
\toprule
Method & GenImage & DRCT-2M & AIGCDetectBenchmark & Avg. \\
\midrule
NPR \textsubscript{\textcolor{blue}{(CVPR'24)}} \cite{tan2024rethinking}  
& 51.5 & 37.3 & 53.1 & 47.3 $\pm$ 7.1 \\
UnivFD \textsubscript{\textcolor{blue}{(CVPR'23)}} \cite{ojha2023towards} 
& 64.1 & 61.8 & 72.5 & 66.1 $\pm$ 4.6 \\
FatFormer \textsubscript{\textcolor{blue}{(CVPR'24)}} \cite{liu2024forgery}  
& 62.8 & 52.2 & 85.0 & 66.7 $\pm$ 13.7 \\
SAFE \textsubscript{\textcolor{blue}{(KDD'25)}} \cite{li2024improving} 
& 50.3 & 59.3 & 50.3 & 53.3 $\pm$ 4.2 \\
C2P-CLIP \textsubscript{\textcolor{blue}{(AAAI'25)}} \cite{tan2024c2p}  
& 74.4 & 59.2 & 81.4 & 71.7 $\pm$ 9.3 \\
AIDE \textsubscript{\textcolor{blue}{(ICLR'25)}} \cite{yan2024sanity} 
& 61.2 & 64.6 & 63.6 & 63.1 $\pm$ 1.4 \\
DRCT \textsubscript{\textcolor{blue}{(ICML'24)}} \cite{chen2024drct}  
& 84.7 & 90.5 & 81.4 & 85.5 $\pm$ 3.8 \\
AlignedForensics \textsubscript{\textcolor{blue}{(ICLR'25)}} \cite{rajan2025aligned} 
& 79.0 & 95.5 & 66.6 & 80.4 $\pm$ 11.8 \\
CO-SPY \textsubscript{\textcolor{blue}{(CVPR'25)}} \cite{cheng2025co} 
& 76.3 & 83.1 & 72.5 & 77.3 $\pm$ 4.4 \\
BFree \textsubscript{\textcolor{blue}{(CVPR'25)}} \cite{guillaro2024bias} 
& \underline{89.5} & \textbf{99.1} & \textbf{88.2} & \underline{92.3} $\pm$ 4.9 \\
\midrule
\rowcolor{myblue}
\textbf{AlignGemini}  
& \textbf{91.7} & \underline{98.1} & \underline{87.7} & \textbf{92.5} $\pm$ 4.3 (\plus{+0.2})\\
\bottomrule
\end{tabular}
\end{adjustbox}
\end{minipage}
\hfill
\begin{minipage}[t]{0.48\textwidth}
\centering
\caption{Comparison with baseline methods under the same training set as \ours.}
\label{tab:ablation_study_backbones}
\vspace{-5pt}
\begin{adjustbox}{width=\linewidth}
\begin{tabular}{lcccccc}
\toprule
Method & Chameleon & WildRF & AIGI-Bench & CO-SPY-Bench/in-the-wild & BFree-Online & Avg. \\
\midrule
UnivFD & 52.0 & 63.2 & 58.2 & 18.9 & 49.8 & 48.4 $\pm$ 15.5 \\
AIDE & 64.2 & 61.3 & 63.2 & 44.2 & 50.2 & 56.6 $\pm$ 8.0 \\
CO-SPY & \underline{66.7} & 70.2 & 65.4 & 52.6 & 53.0 & 61.6 $\pm$ 7.3 \\
NPR & 58.3 & 56.9 & 54.2 & 12.3 & 48.9 & 46.1 $\pm$ 17.2\\
SAFE & 61.2 & 62.4 & 55.3 & 21.7 & 52.6 & 50.6 $\pm$ 14.9 \\
DINOv2-ViT-L-14 & 51.7 & 57.8 & 57.8 & 13.4 & 49.2 & 46.0 $\pm$ 16.6 \\
Qwen2.5-VL-7B & {65.7} & \underline{86.3} & \underline{80.4} & \underline{67.6} & \underline{82.9} & \underline{76.6} $\pm$ 8.3 \\
\midrule
\rowcolor{myblue}
\textbf{AlignGemini} & \textbf{85.5} & \textbf{92.9} & \textbf{88.8} & \textbf{94.3} & \textbf{97.5} & \textbf{91.8} $\pm$ 4.2 (\plus{+15.2}) \\
\bottomrule
\end{tabular}
\end{adjustbox}
\end{minipage}
\vspace{-5pt}
\end{table*}

\vspace{-5pt}
\paragraph{Evaluation Metrics and Comparative Methods.}
Balanced accuracy is adopted as the primary metric, defined as the mean of the accuracies on real and synthetic images. The competitive methods include 
NPR~\cite{tan2024rethinking}, FatFormer~\cite{liu2024forgery}, UnivFD~\cite{ojha2023towards}, SAFE~\cite{li2024improving}, AIDE~\cite{yan2024sanity}, C2P-CLIP~\cite{tan2024c2p}, DRCT~\cite{chen2024drct}, AlignedForensics~\cite{rajan2025aligned}, CO-SPY~\cite{cheng2025co}, and BFree~\cite{guillaro2024bias}, and Forensic-Chat~\cite{lin2025seeing}.

All baselines are evaluated using their officially released checkpoints, without per\mbox{-}benchmark retuning.
For AlignGemini, we consistently adopt a fine-tuned Qwen2.5-VL-7B as the VLM branch and a fine-tuned DINOv2 as the expert branch across all benchmarks.
We organize our experiments around five key questions:

\vspace{5pt}
\noindent\colorbox{gray!20}{%
  \parbox{\dimexpr\linewidth-2\fboxsep\relax}{%
    \small \textbf{Q1.} Does AlignGemini achieve stronger generalization? 
  }%
}

\begin{figure*}[t!]
    \centering
    \begin{adjustbox}{width=0.9\textwidth}
        \begin{subfigure}[b]{0.24\textwidth}
            \centering
            \includegraphics[width=\textwidth]{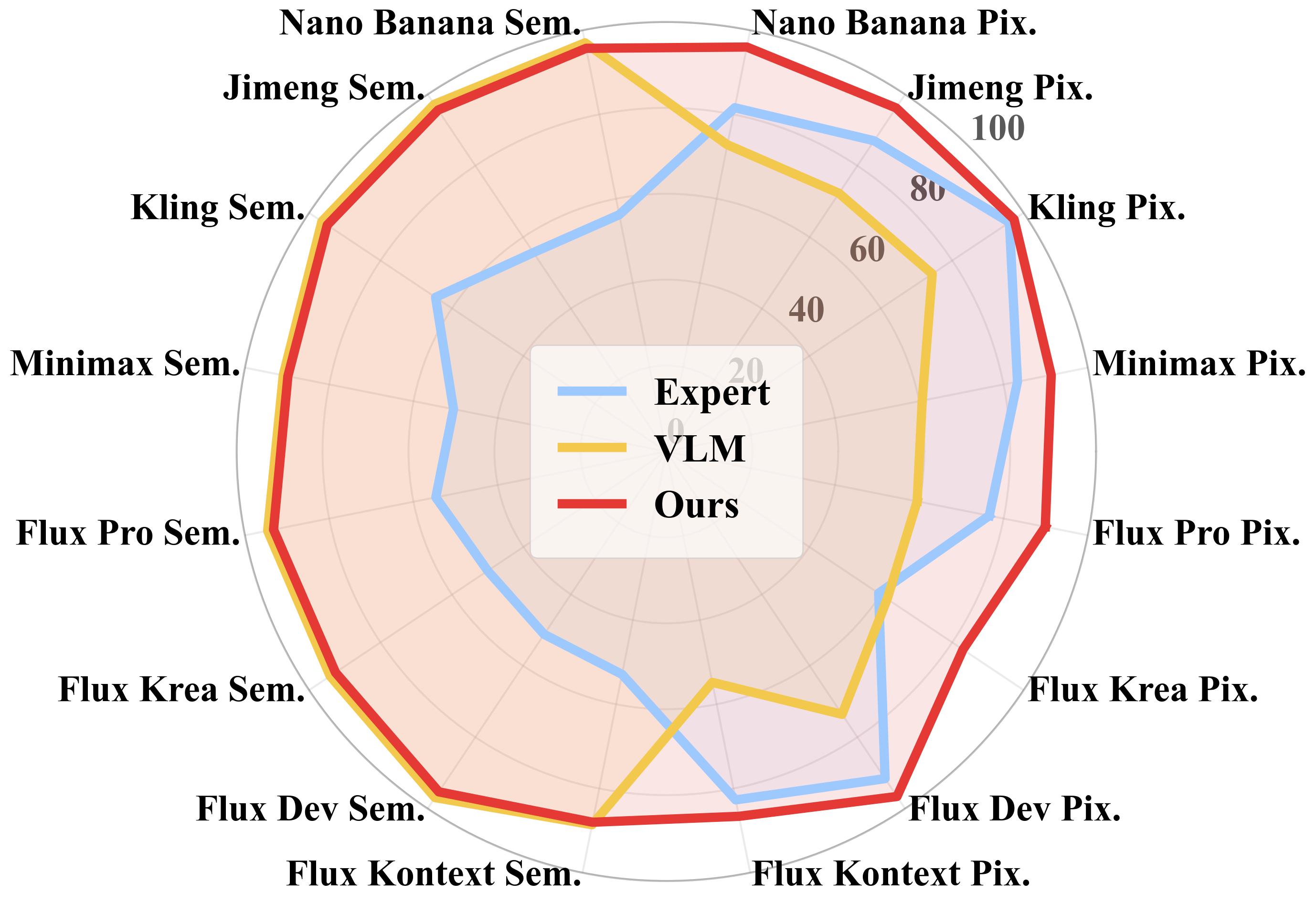}
            \caption{VLM (pure) + Expert (pure)}
        \end{subfigure}
        \hfill 
        \begin{subfigure}[b]{0.24\textwidth}
            \centering
            \includegraphics[width=\textwidth]{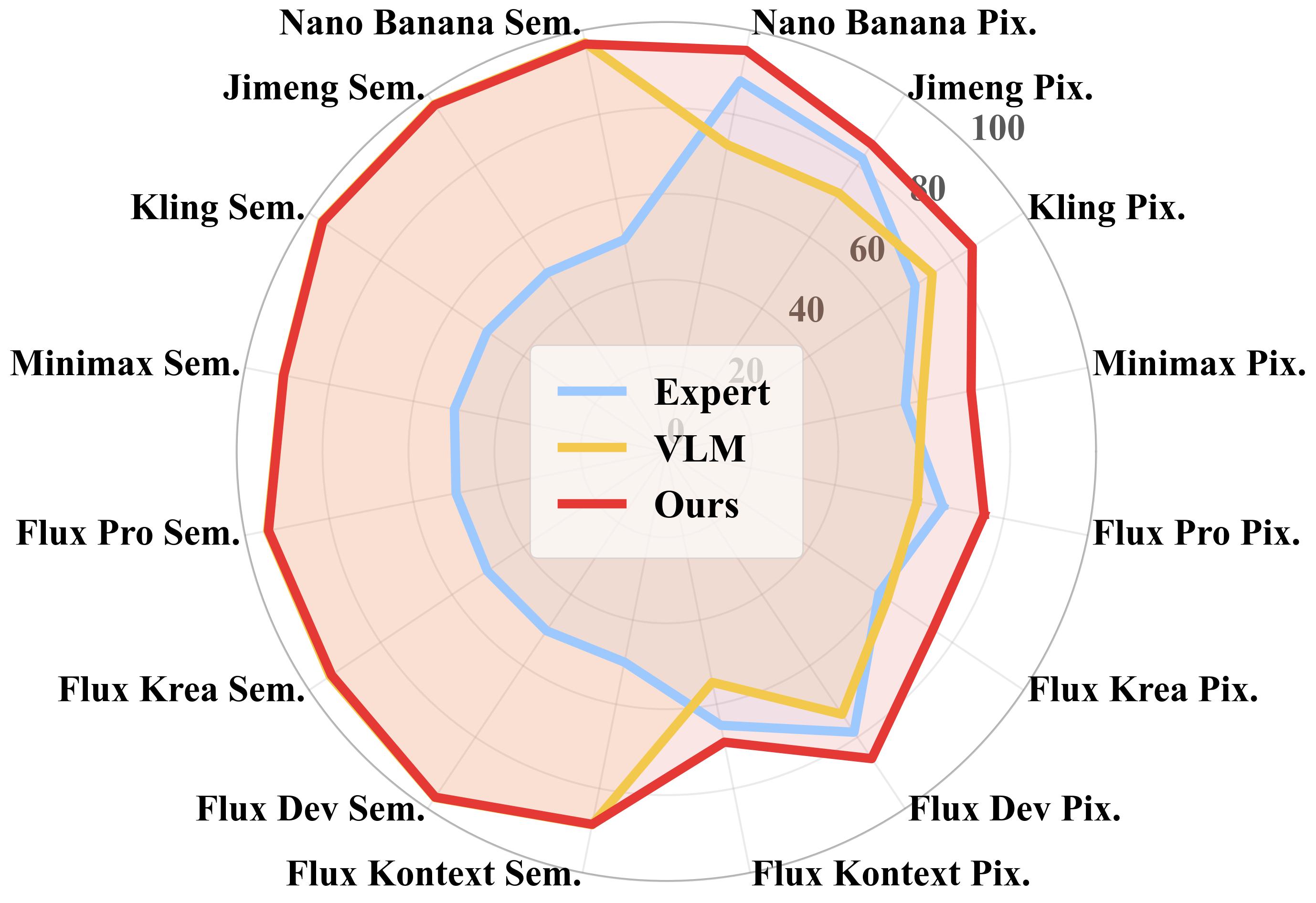}
            \caption{VLM (pure) + Expert (blend)}
        \end{subfigure}
        \hfill
        \begin{subfigure}[b]{0.24\textwidth}
            \centering
            \includegraphics[width=\textwidth]{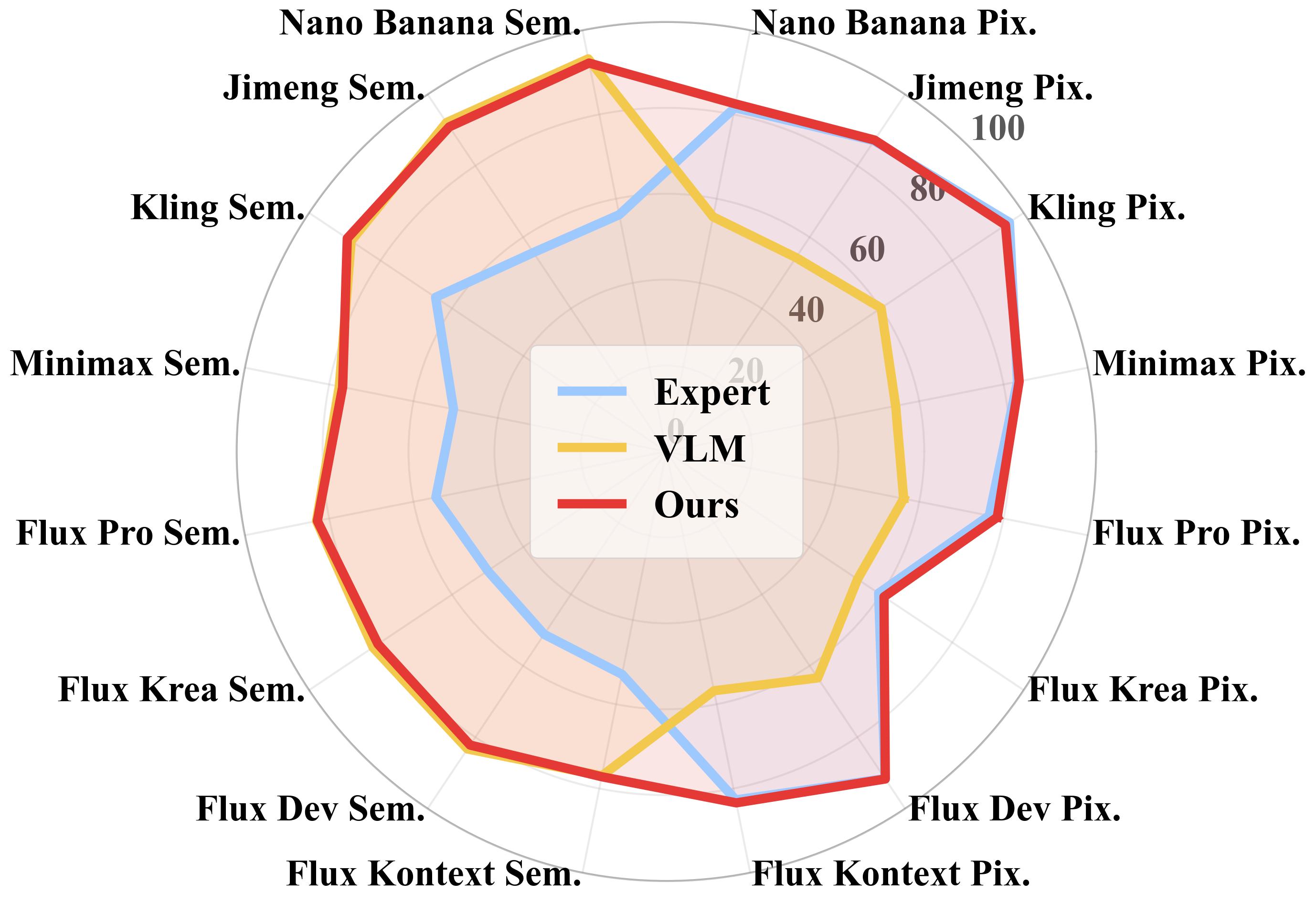}
            \caption{VLM (blend) + Expert (pure)}
        \end{subfigure}
        \hfill
        \begin{subfigure}[b]{0.24\textwidth}
            \centering
            \includegraphics[width=\textwidth]{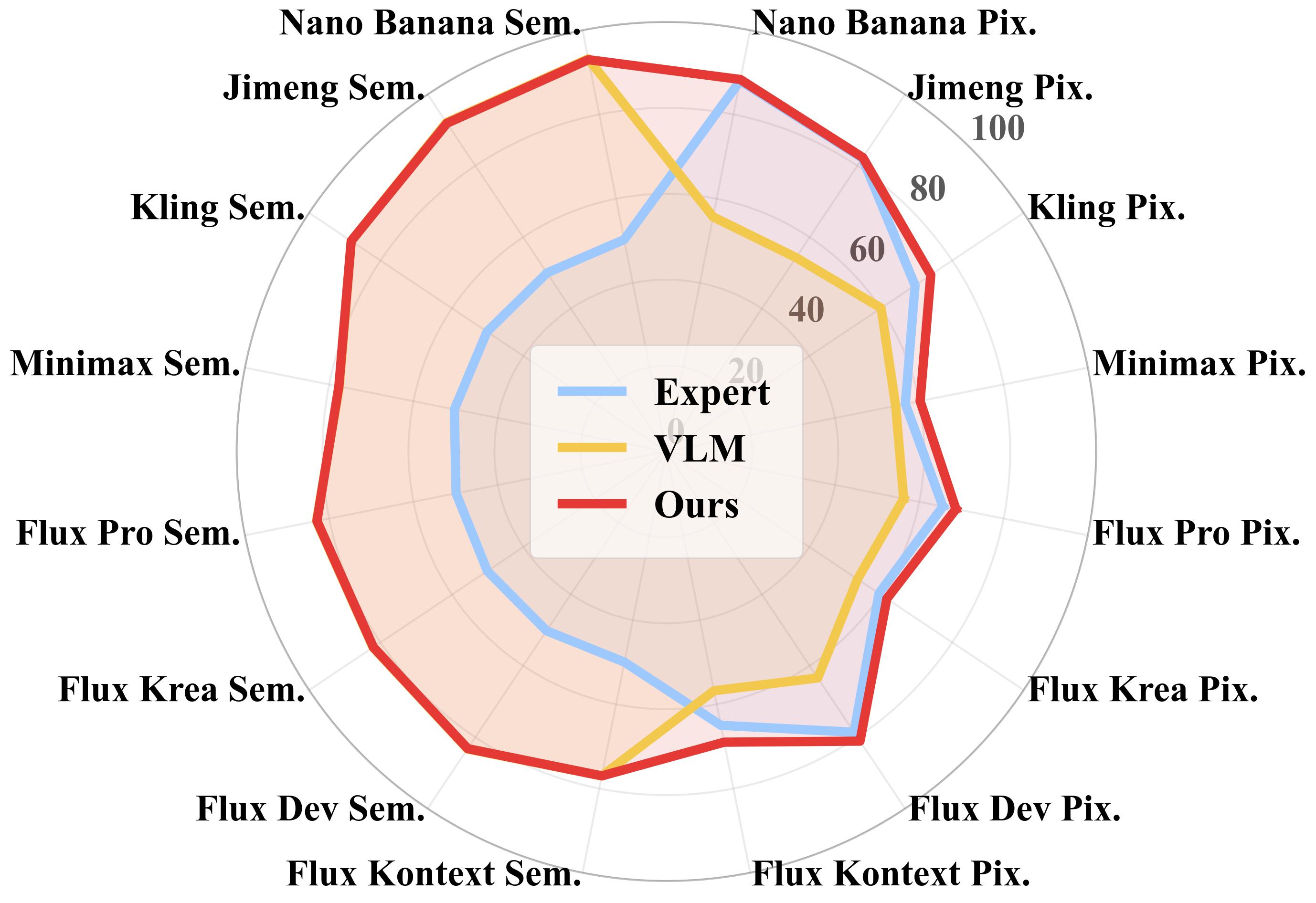}
            \caption{VLM (blend) + Expert (blend)}
        \end{subfigure}
    \end{adjustbox}
    \vspace{-5pt}
     \caption{Ablation on the impact of supervision disentanglement.
``Pure'' denotes semantic-only supervision for the VLM branch and artifact-only supervision for the Expert branch, while ``blend'' applies combined semantic and pixel-level supervision.
Pure supervision encourages stronger branch specialization and more effective complementarity, whereas blended supervision dilutes specialization and degrades detection performance.}
    \label{fig:ablation-pularity}
\end{figure*}

\begin{figure*}[t!]
    \centering
    \begin{adjustbox}{width=0.95\textwidth}
        \begin{subfigure}[b]{0.19\textwidth}
            \centering
            \includegraphics[width=\textwidth]{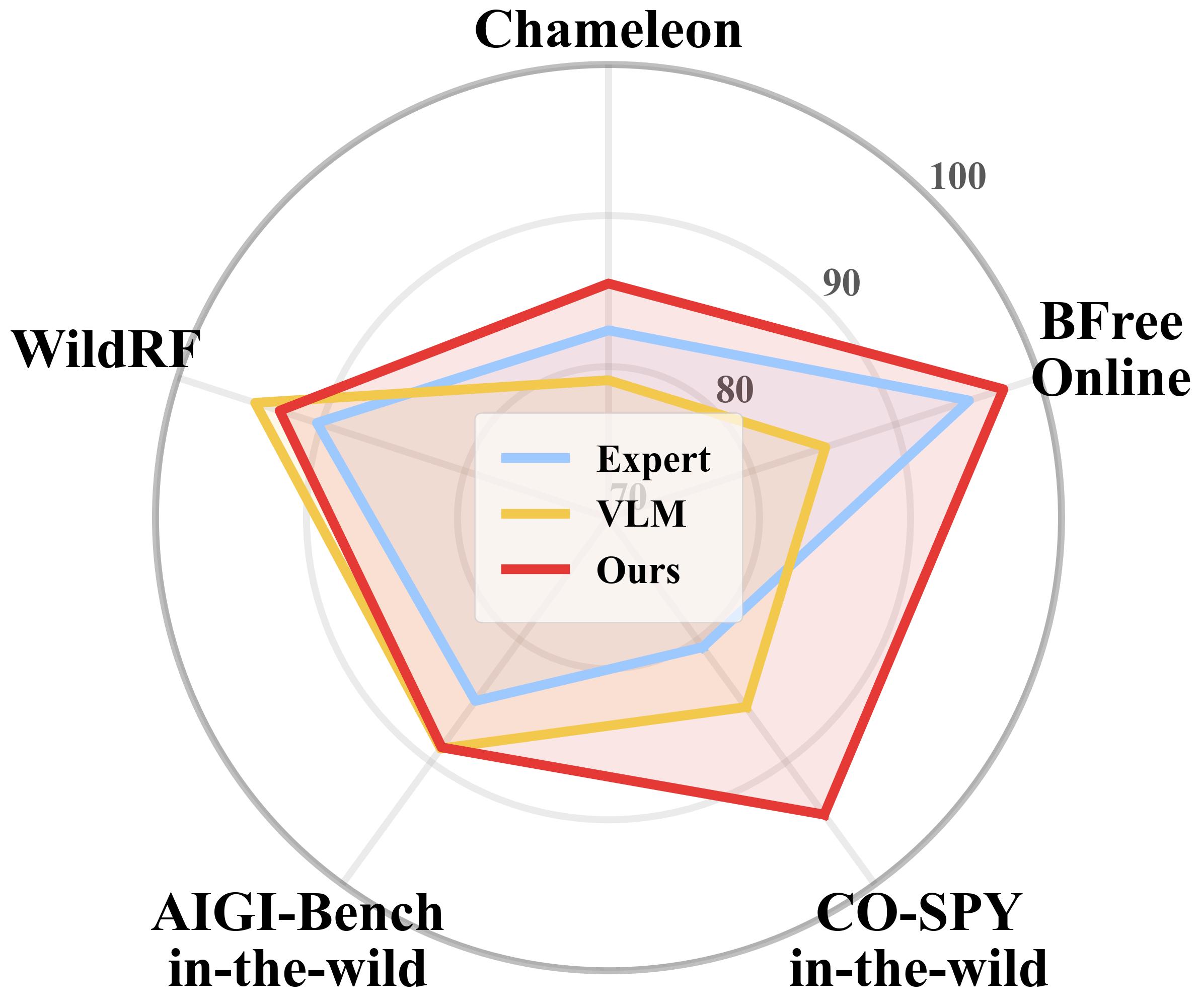}
            \caption{AlignGemini}
        \end{subfigure}
        \hfill
        \begin{subfigure}[b]{0.19\textwidth}
            \centering
            \includegraphics[width=\textwidth]{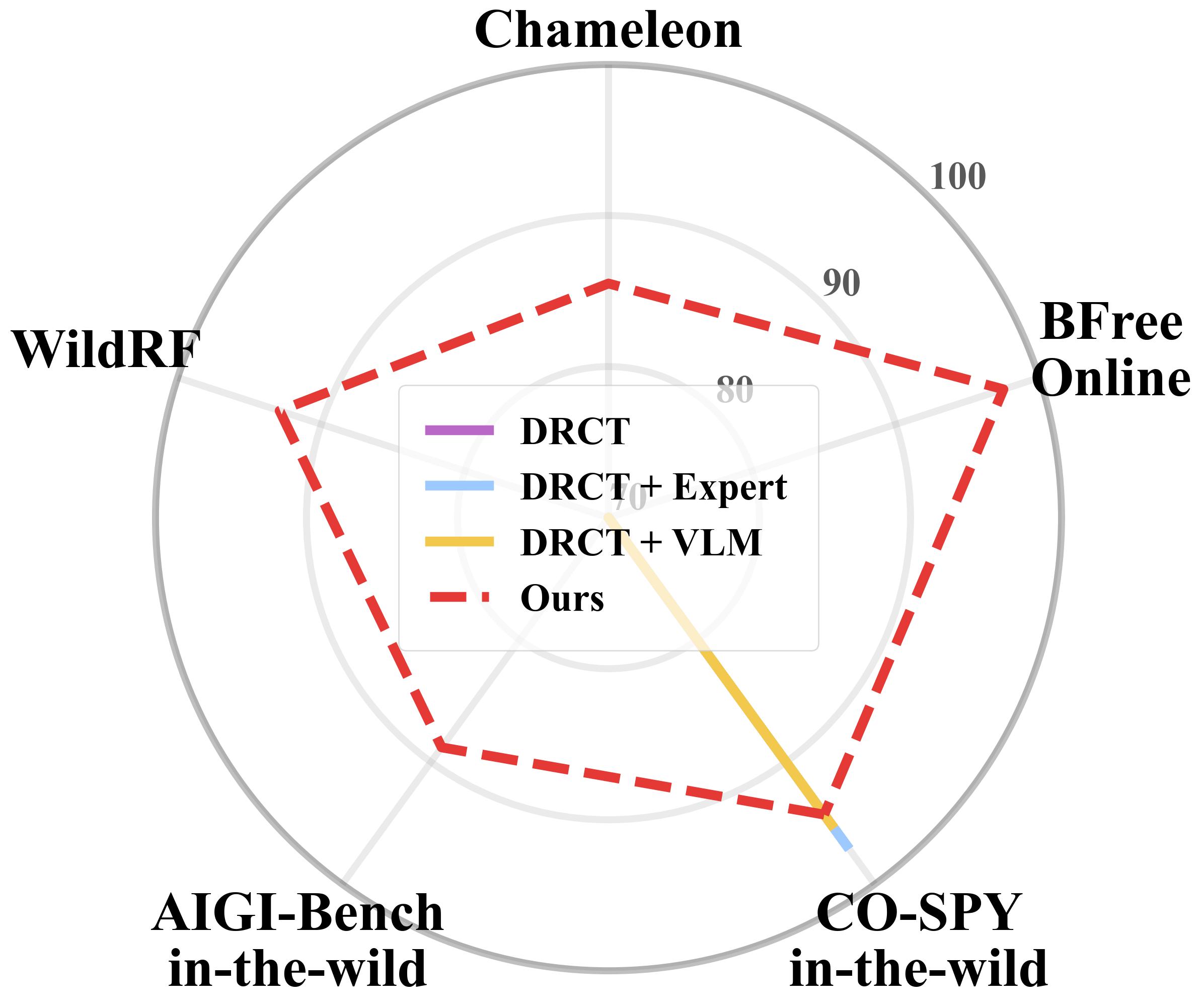}
            \caption{DRCT}
        \end{subfigure}
        \hfill
        \begin{subfigure}[b]{0.19\textwidth}
            \centering
            \includegraphics[width=\textwidth]{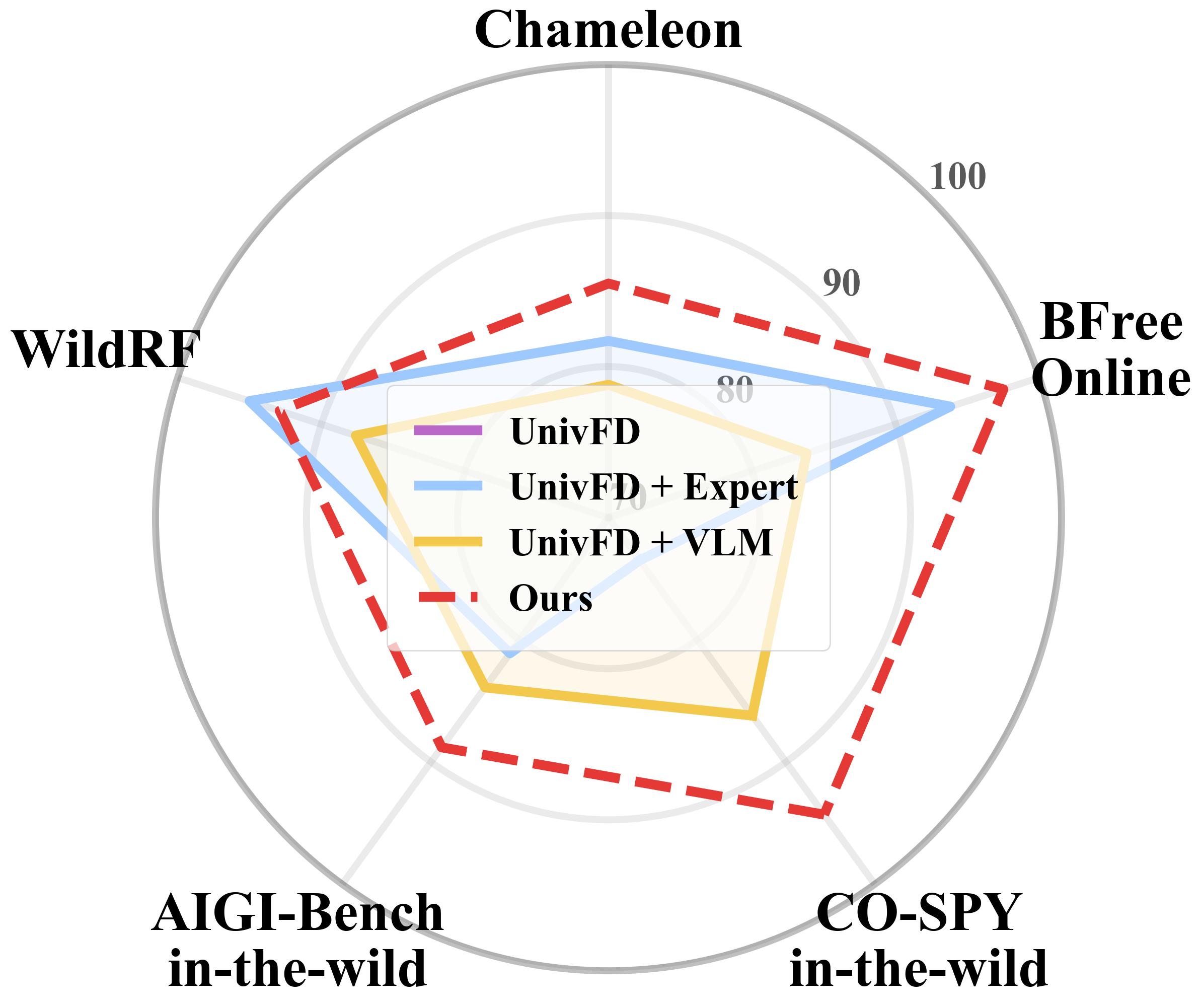}
            \caption{UnivFD}
        \end{subfigure}
        \hfill
        \begin{subfigure}[b]{0.19\textwidth}
            \centering
            \includegraphics[width=\textwidth]{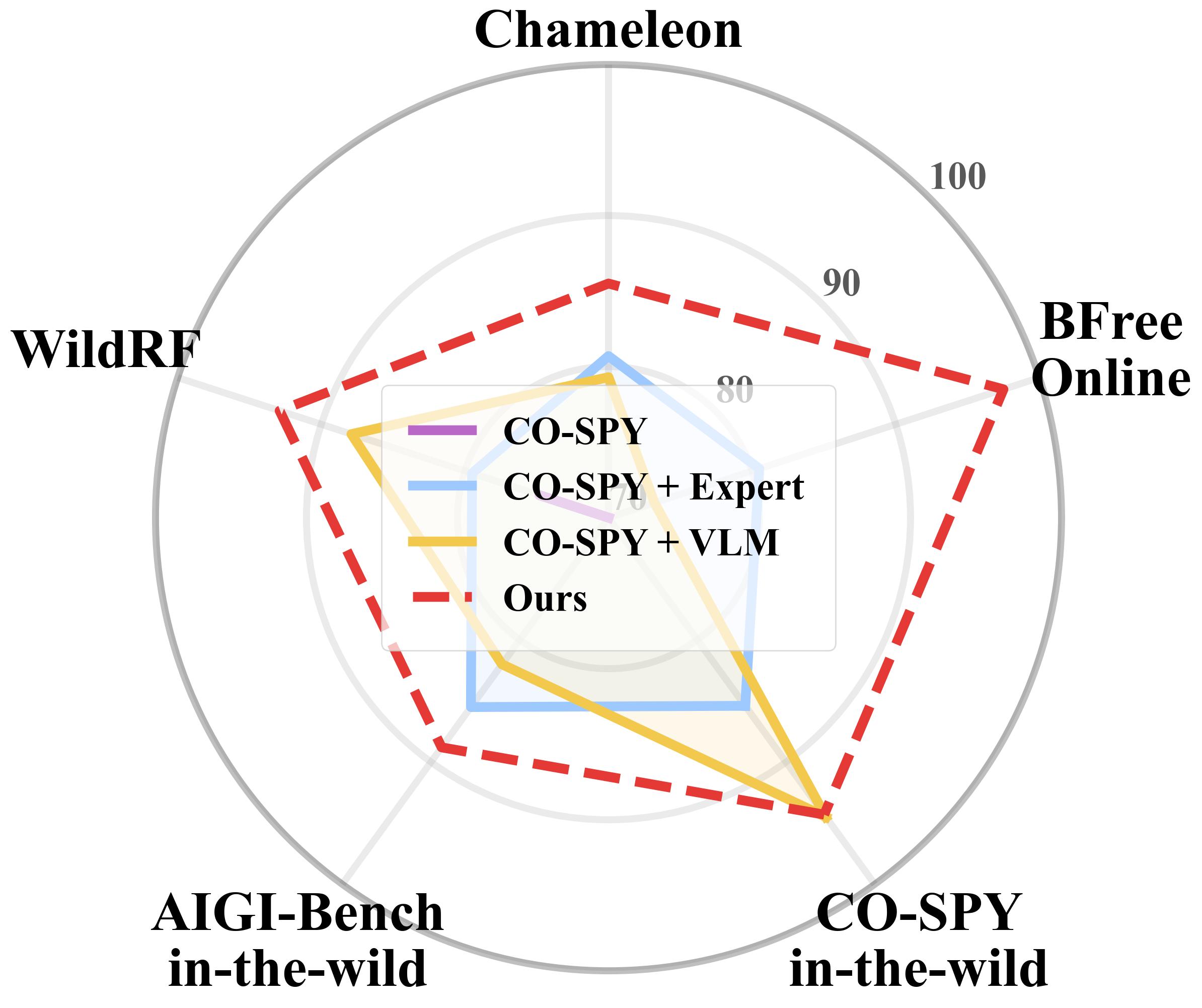}
            \caption{CO-SPY}
        \end{subfigure}
        \hfill 
        \begin{subfigure}[b]{0.19\textwidth}
            \centering
            \includegraphics[width=\textwidth]{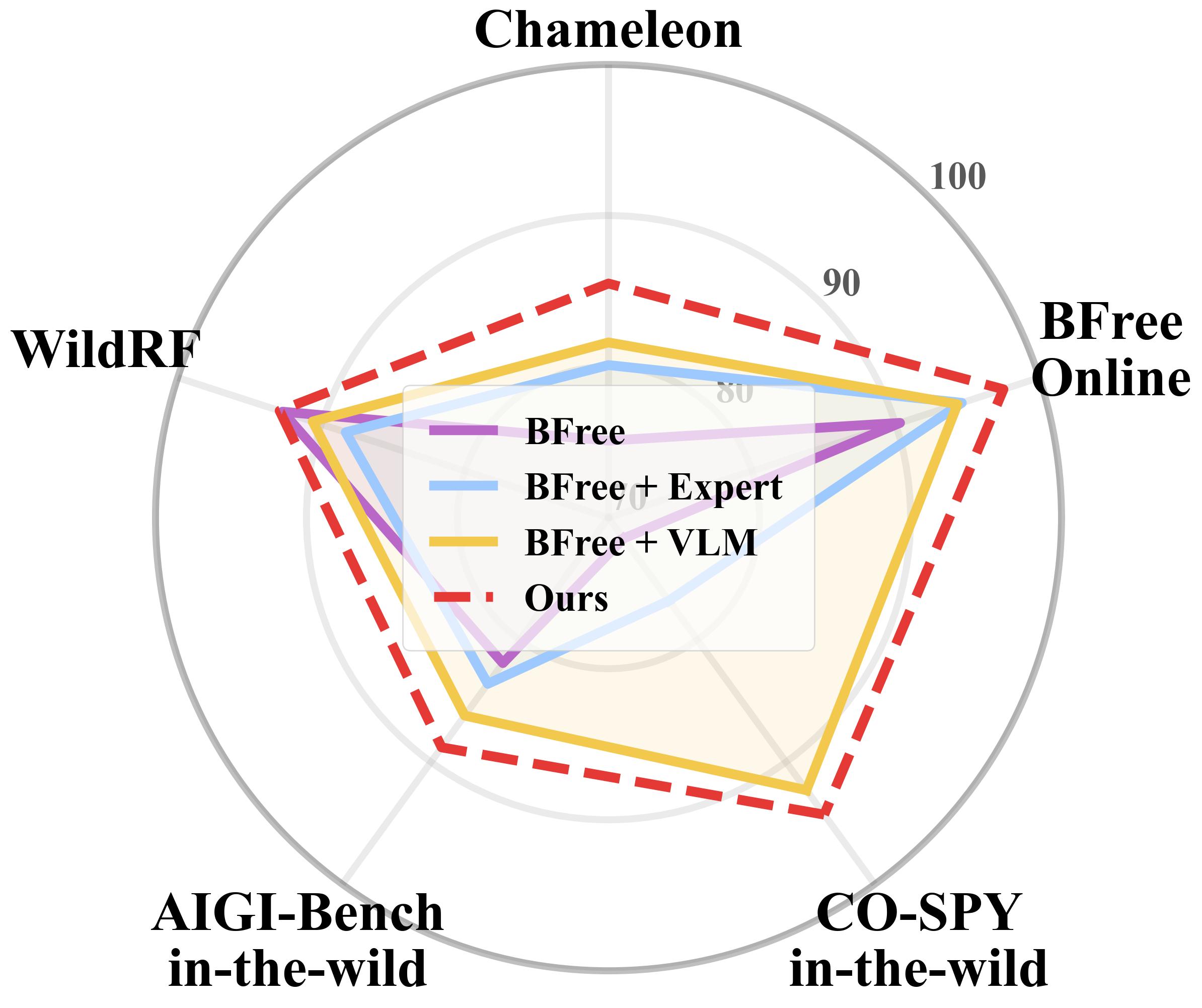}
            \caption{BFree}
        \end{subfigure}
    \end{adjustbox}
    \vspace{-10pt}
    \caption{Ablation on the effect of using two models across five in-the-wild benchmarks. 
    Each subfigure reports the in-the-wild performance of a baseline method augmented with one branch of AlignGemini, resulting in capacity-matched two-model–to–two-model comparisons.
    \textbf{AlignGemini consistently achieves non-trivial performance improvements, indicating that its gains are not solely attributable to the use of multiple models.}} 
    \label{fig:ablation-two-models}
\end{figure*}

\paragraph{Comparison on Five In-the-Wild Benchmarks.}
Table~\ref{tab:compare-in-the-wild} reports performance on five in-the-wild benchmarks.
In this challenging regime, most baselines achieve accuracies below 70, underscoring the difficulty of real-world generalization. AlignGemini consistently surpasses all baselines, improving accuracy by \textbf{+9.5} and demonstrating its superior generalization.

\paragraph{Comparison on AIGI-Now and Three Self-Synthesized Benchmarks.}
Table~\ref{tab:compare-aiginow} and Table~\ref{tab:compare-self-synthesized} compare detectors on AIGI-Now, GenImage, DRCT-2M, and AIGCDetectBenchmark.
Results on AIGI-Now reveal that most existing detectors exhibit limited pixel-level generalization to modern image generators: only DRCT and CO-SPY achieve balanced accuracies above 80\%.
In contrast, AlignGemini improves pixel-level accuracy by \textbf{+10.7}, demonstrating substantially stronger generalization.
For semantic detection, nearly all baselines fail to identify semantic inconsistencies, yielding average accuracies below 70\%, whereas AlignGemini achieves a markedly higher accuracy of \textbf{93.1}.
AlignGemini also achieves the highest average accuracy across three self-synthesized benchmarks.
Overall, these results indicate that prior detectors are largely insensitive to semantic cues, while AlignGemini achieves strong generalization across both semantic and pixel-level subtasks.

\vspace{5pt}
\noindent\colorbox{gray!20}{%
  \parbox{\dimexpr\linewidth-2\fboxsep\relax}{%
    \small \textbf{Q2.} Where do AlignGemini's performance gains come from?
  }%
}

\paragraph{Gains from Model-Supervision Alignment.}
To isolate the effect of orthogonal supervision disentanglement, we retrain all baselines on identical corpora comprising the same semantic and pixel-artifact supervision sets.
As shown in Table~\ref{tab:ablation_study_backbones}, AlignGemini achieves a substantial gain of \textbf{+15.2\%} over the best-performing baseline under this controlled setting.
This improvement stems from model-supervision alignment, which trains each branch exclusively on inductively matched data.

\paragraph{Gains from Supervision Disentanglement.}
Figure~\ref{fig:ablation-pularity} further isolates this effect by comparing pure against blended supervision for each branch.
The results show that disentangled supervision markedly enhances each branch's specialized discriminative capability, thereby improving overall performance.
When blended supervision is substituted, a branch's inductive strength is diluted; although minor gains appear on off-target capabilities, overall performance decreases substantially.
Together, these results confirm the effectiveness of both model-supervision alignment and supervision disentanglement as sources of AlignGemini's gains.

\vspace{-5pt}

\paragraph{Gains Are Not Attributable to Using Two Models.}
Figure~\ref{fig:ablation-two-models} presents a two-model--to--two-model comparison, ruling out the possibility that simply adding a second branch to any baseline could achieve the same generalization as AlignGemini.
Across all five in-the-wild benchmarks, AlignGemini's two-branch integration yields consistently higher accuracy.
These results indicate that the gains are not attributable to merely employing multiple models—since augmenting a baseline with either branch alone fails to achieve comparable performance—but instead arise from model-supervision alignment and supervision disentanglement that maximize specialized discriminative capability and promote complementarity.

\vspace{5pt}
\noindent\colorbox{gray!20}{%
  \parbox{\dimexpr\linewidth-2\fboxsep\relax}{%
    \small \textbf{Q3.} Does AlignGemini exhibit higher robustness?
  }%
} 
\vspace{-5pt}

\begin{figure*}[tb!]
    \centering
    \includegraphics[width=0.9\textwidth]{fig/jpg/robust_legend.png}
    \begin{subfigure}[b]{0.2\textwidth}
        \centering
        \includegraphics[width=\textwidth]{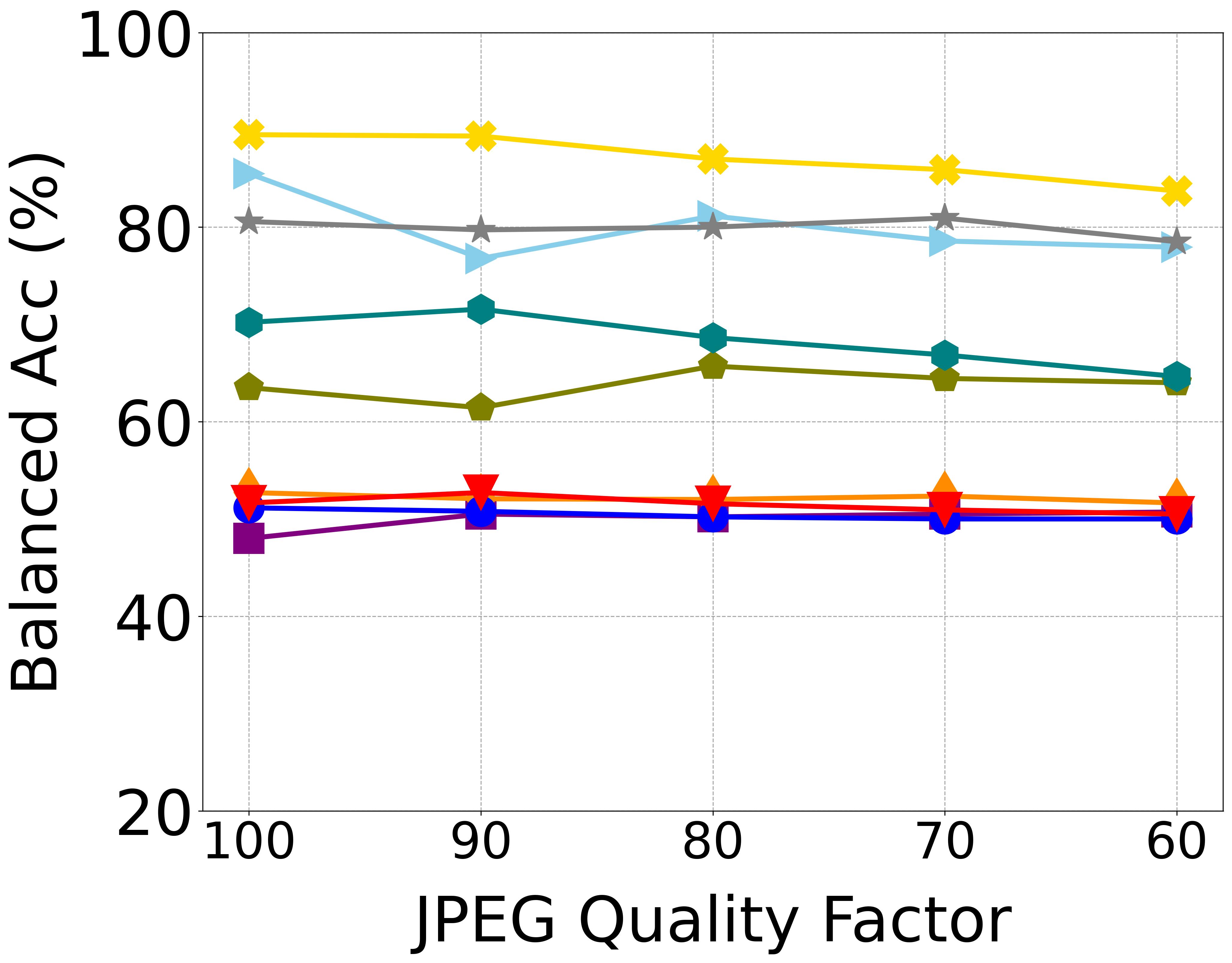}
        \caption{\small JPEG compression}
    \end{subfigure}%
    \hfill
    \begin{subfigure}[b]{0.2\textwidth}
        \centering
        \includegraphics[width=\textwidth]{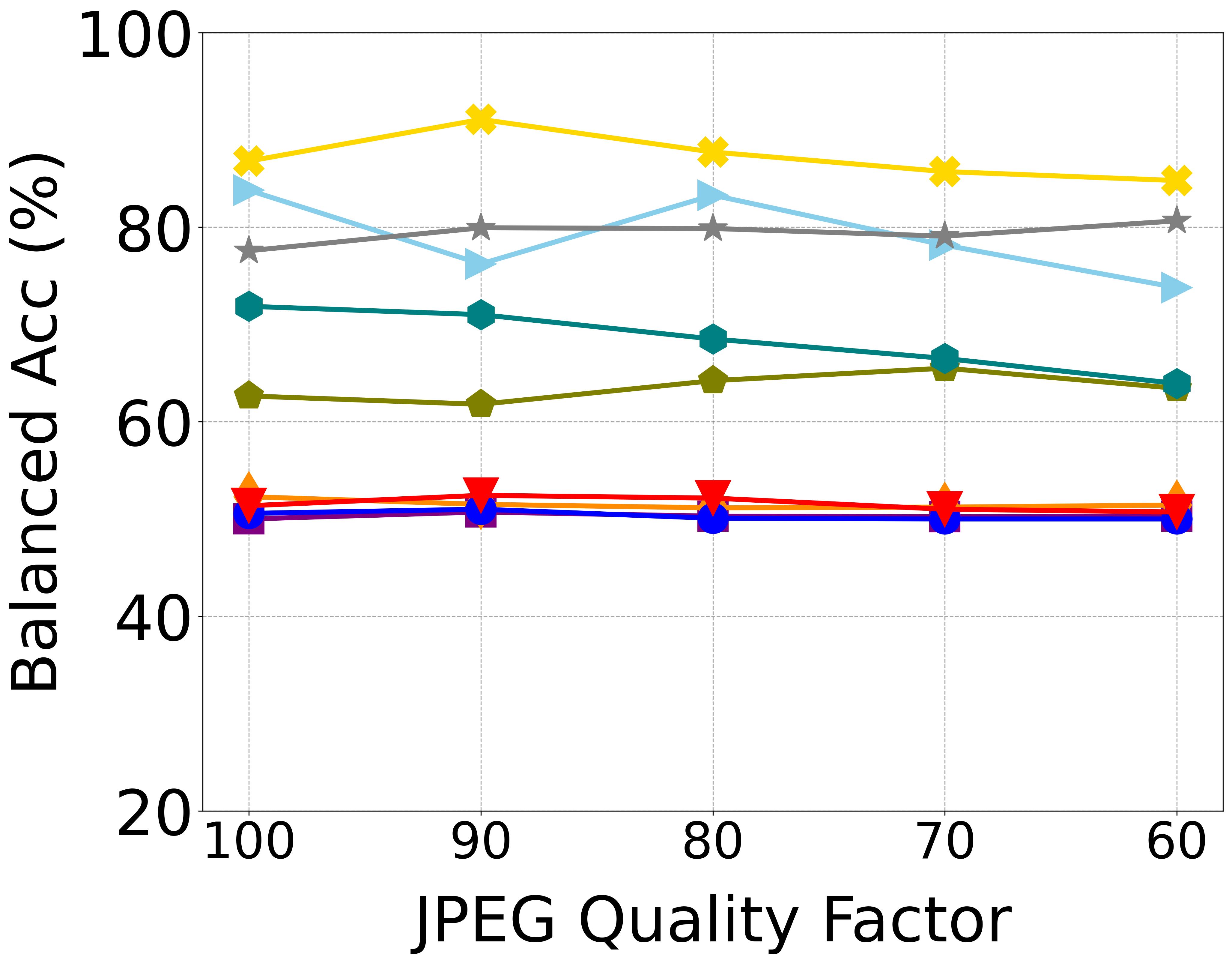}
        \caption{\small Double JPEG}
    \end{subfigure}%
    \hfill
    \begin{subfigure}[b]{0.2\textwidth}
        \centering
        \includegraphics[width=\textwidth]{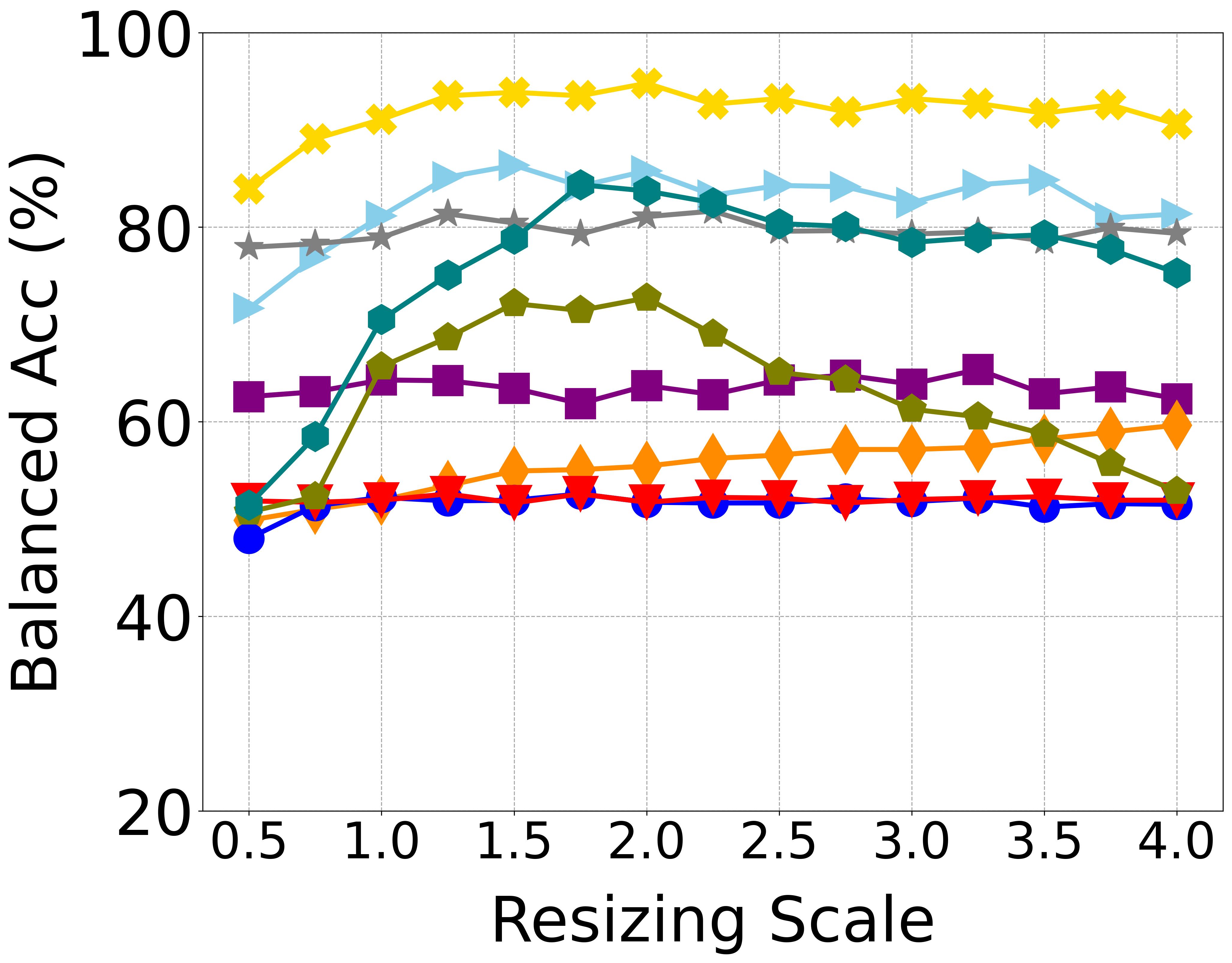}
        \caption{\small Resizing}
    \end{subfigure}
    \hfill
    \begin{subfigure}[b]{0.2\textwidth}
        \centering
        \includegraphics[width=\textwidth]{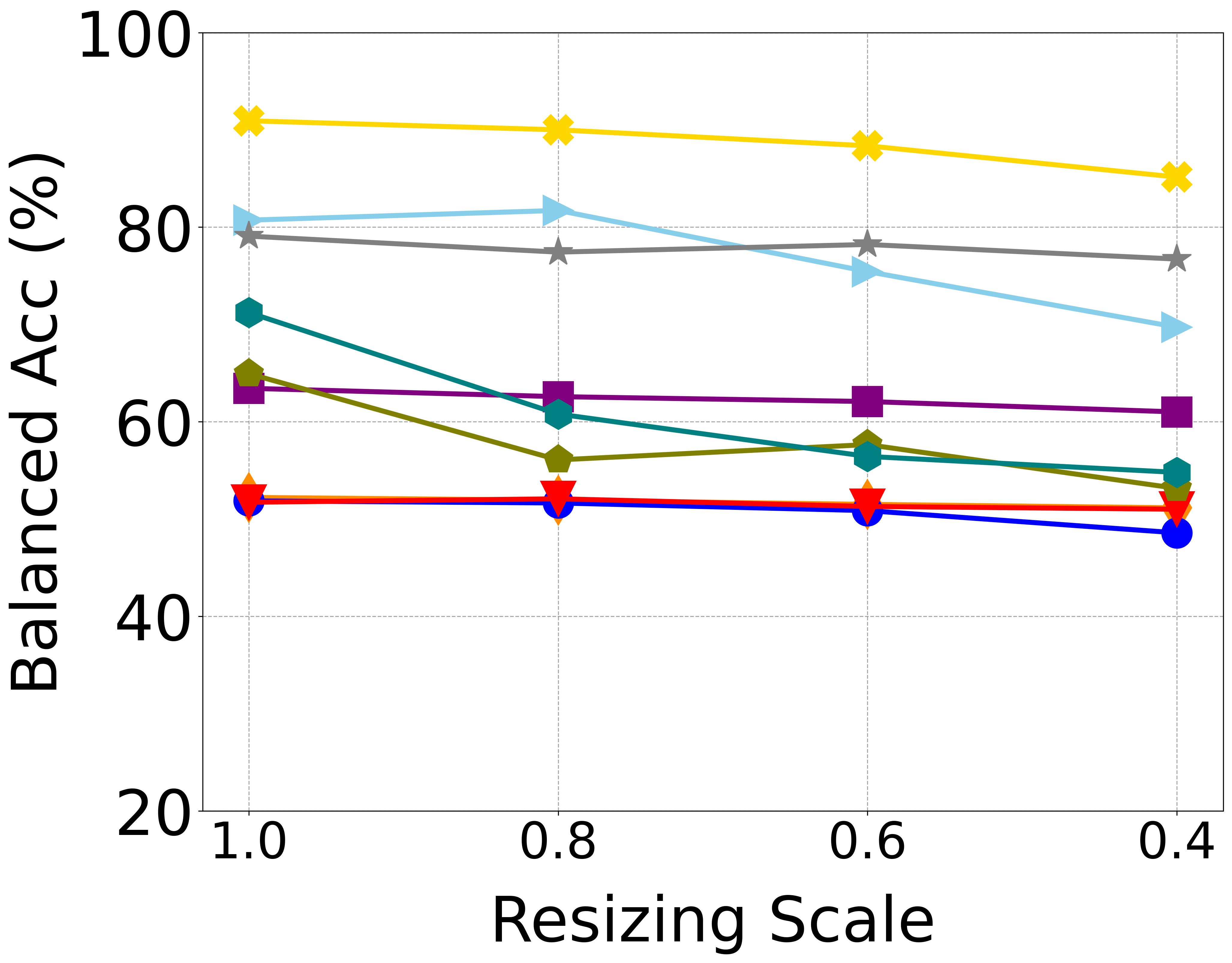}
        \caption{\small Double Resizing}
    \end{subfigure}
    \caption{Robustness evaluation on AIGI-Now (pixel), which assesses robustness under common post-processing operations.
        Double JPEG compression applies the same JPEG quality factor twice in succession. 
        Double resizing downsamples the image and then upsamples it back to the original resolution. 
        AlignGemini demonstrates superior robustness across all post-processing.} 
    \label{fig:robustness}
\end{figure*}

\begin{figure*}[h]
\centering
\includegraphics[width=.95\textwidth]{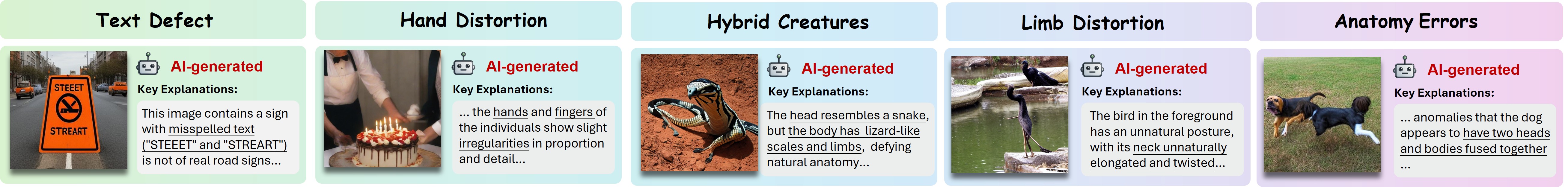} \vspace{-5pt}
\caption{The VLM branch of AlignGemini discriminates fine-grained semantic flaws despite being trained exclusively on obviously inconsistent images, suggesting that pure semantic supervision activates the VLM's internal knowledge rather than teaching new discriminative patterns.}
\label{fig:fine-grained} 
\end{figure*}

\begin{table*}[t!]
\centering
\caption{Evaluate the in-the-wild accuracies and FLUX-specific accuracies of AlignGemini when upgrading the VLM backbone and the pixel experts. \textbf{SD}/\textbf{FLUX} denote experts trained on SD2/FLUX VAE-reconstructed images, respectively. \textbf{SD+FLUX} uses both experts jointly. The results show that performance consistently improves with stronger backbones and additional experts.}
\vspace{-5pt}
\label{tab:extensible}

\begin{adjustbox}{width=0.95\linewidth}
\begin{tabular}{lll c c c c c c ccccc}
\toprule
\multicolumn{3}{c}{Method} 
& \multicolumn{6}{c}{\textbf{In-the-wild benchmarks}} 
& \multicolumn{5}{c}{\textbf{AIGI-Now}} \\

\cmidrule(lr){4-9} \cmidrule(lr){10-14}
& & 
& {Chameleon} 
& {WildRF} 
& {AIGI-Bench} 
& {CO-SPY-Bench/in-the-wild} 
& {BFree-Online} 
& {Avg.} 
& {Flux Pro.}
& {Flux Krea.}
& {Flux Dev}
& {Flux Kontext.}
& {Avg.}
\\

\midrule

\multirow{5}{*}{Baseline}
& Qwen2.5-VL-7B (Untrained)
& -
& 59.1 
& 83.8
& 75.7
& 57.4
& 77.6
& 70.7 $\pm$ 10.5
& 69.6
& 67.7
& 73.3
& 66.9
& 69.4 $\pm$ 2.5
\\

& BFree
& - 
& {75.1} 
& {92.7} 
& {81.9} 
& 71.7 
& {90.3}  & 82.3 $\pm$ 8.2
& 63.1 & 55.5 & 64.6 & 65.3 & 62.1 $\pm$ 3.9\\

& CO-SPY
& - 
& 68.8 
& 74.9 
& 66.5 
& 64.6 
& 55.2 & 66.0 $\pm$ 6.4
& 75.0 & 81.5 & 81.7 & 65.6 & 76.0 $\pm$ 6.6
\\

& CLIP ViT-L/14
& SD
& 82.5
& 90.8
& 85.0
& 73.7
& 95.1 & 85.4 $\pm$ 7.3
& 75.8 & 76.3 & 96.4 & 83.9 & 83.1 $\pm$ 8.3
\\

& SigLIP2
& SD
& 81.9
& 90.6
& 86.2
& 76.6
& 95.0 & 86.1 $\pm$ 6.4
& 76.6 & 72.2 & 96.2 & 83.4 & 82.1 $\pm$ 9.1
\\

\midrule
\rowcolor{myblue}
& Qwen2.5-VL-7B & SD
& 85.5 
& 92.9 
& 88.8 
& 94.3 
& 97.5 
& 91.8 $\pm$ 4.2 (\plus{+5.7})
& 91.5 & 87.7 & 95.9 & 87.3 
& 90.6 $\pm$ 3.5 (\plus{+7.5})\\

\rowcolor{myblue}
& Qwen2.5-VL-7B & SD + FLUX
& 85.6 
& 93.1 
& 88.9 
& 94.3 
& 97.5 & 91.9 $\pm$ 4.2 (\plus{+5.8})
& 91.5 & \underline{92.1} & 96.5 & \underline{89.2} 
& \underline{92.3} $\pm$ 2.6 (\plus{+9.2})\\

\rowcolor{myblue}
& Qwen3-VL-32B & SD
& 87.9 
& 93.9 
& 90.4 
& {96.0} 
& 95.3 & 92.7 $\pm$ 3.1 (\plus{+6.6})
& 89.7 & 83.6 & 95.5 & 87.1 
& 88.9 $\pm$ 4.3 (\plus{+5.9}) \\

\rowcolor{myblue}
\textbf{AlignGemini}
& Qwen3-VL-32B & SD + FLUX
& 88.4 
& \underline{94.4} 
& {90.8} 
& {96.3} 
& 95.6 
& 93.1 $\pm$ 3.0 (\plus{+7.0})
& 89.8 & 90.5 & 96.4 & 89.1 
& 91.4 $\pm$ 2.9 (\plus{+8.4}) \\

\rowcolor{myblue}
& Qwen3-VL-235B & SD
& \underline{89.8} 
& 94.0 
& \underline{91.2} 
& \underline{97.2} 
& \underline{97.8} & \underline{94.0} $\pm$ 3.2 (\plus{+7.9})
& \underline{93.1} & 89.9 & \underline{97.0} & 87.8 
& 92.0 $\pm$ 3.5 (\plus{+8.9}) \\

\rowcolor{myblue}
& Qwen3-VL-235B & SD + FLUX
& \textbf{90.3} 
& \textbf{94.5} 
& \textbf{91.6} 
& \textbf{97.5} 
& \textbf{98.2} & \textbf{94.4} $\pm$ 3.1 (\plus{+8.3})
& \textbf{93.2} & \textbf{94.0} & \textbf{97.6} & \textbf{89.8} 
& \textbf{93.7} $\pm$ 2.8 (\plus{+10.6}) \\
\bottomrule
\end{tabular}
\end{adjustbox}
\end{table*}

\paragraph{Comparison on Robustness.}
Figure~\ref{fig:robustness} evaluates robustness under four post-processing operations.
Pixel-artifact detectors degrade noticeably, especially under double resizing at scale factor 0.4, whereas detectors incorporating semantic features remain more stable. AlignGemini performs best across all settings.

\vspace{5pt}
\noindent\colorbox{gray!20}{%
  \parbox{\dimexpr\linewidth-2\fboxsep\relax}{%
    \small \textbf{Q4.} Is the VLM trained on obvious semantic inconsistencies still sensitive to fine-grained semantic flaws?
  }%
} 

\paragraph{Pure Supervision Activates the VLM's Inner Knowledge.}
Figure~\ref{fig:fine-grained} shows cases where the VLM branch of AlignGemini discriminates fine-grained semantic flaws, despite being trained exclusively on obviously inconsistent images.
We attribute this to pure semantic supervision activating the VLM's internal knowledge of normality and anomaly rather than teaching it new discriminative patterns.
Moreover, this semantic discriminative capability scales with model size and can be enhanced by adopting larger backbones.

\vspace{5pt}
\noindent\colorbox{gray!20}{%
  \parbox{\dimexpr\linewidth-2\fboxsep\relax}{%
    \small \textbf{Q5.} Is AlignGemini extensible?
  }%
}

\paragraph{Extensibility of AlignGemini. }
We emphasize that, our core contribution is a general design principle rather than a specific detector instantiation.
Accordingly, AlignGemini is intentionally implemented as a lightweight proof of concept, with simplified data construction and fine-tuning strategies.
Even under this constrained setting, AlignGemini exhibits improved generalization when scaled with stronger backbones or additional training data.
As shown in Table~\ref{tab:extensible}, we examine two complementary extension directions for AlignGemini.
\textbf{VLM branch:} Scaling the backbone from Qwen2.5-VL-7B to Qwen3-VL-235B improves accuracy on Chameleon from 85.5 to 90.3, indicating that stronger VLM backbones offer an effective scaling path.
\textbf{Expert branch:} Adding an additional expert trained on FLUX VAE reconstructions further enhances performance, particularly on FLUX-generated images.
Overall, these results demonstrate that AlignGemini is readily extensible and that its performance scales naturally along both branches.

 \vspace{-5pt}
\section{Conclusion}
In this paper, we identify supervision design as a previously overlooked yet critical factor for generalization in AIGI detection.
Through controlled experiments, we show that aligning each model branch with inductively matched supervision—rather than scaling data or architectures—unlocks substantially stronger generalization.
Extensive experiments further confirm the improved generalization, robustness, and extensibility of our approach.
In addition, we release \textbf{AIGI-Now}, the first benchmark designed for disentangled evaluation of semantic and pixel-level discrimination capabilities.

\vspace{-5pt} 
\begin{acks} 
This work was supported by the
\grantsponsor{NSFC}{National Natural Science Foundation of China}
{https://doi.org/10.13039/501100001809}
(Grant Nos.~\grantnum{NSFC}{62272164},
\grantnum{NSFC}{62572188}, and
\grantnum{NSFC}{62306113}). Supported by the Open Research Project of the
\grantsponsor{SKLICT}
{State Key Laboratory of Industrial Control Technology, China}{}
(Grant No.~\grantnum{SKLICT}{ICT2026A25}).
\end{acks}

\bibliographystyle{ACM-Reference-Format}
\bibliography{ref}

\appendix

In this supplementary material, we provide additional technical and evaluative details:
\begin{itemize}

\item Section~\ref{sec:detailed_results} presents \textbf{additional comparison results}: the detailed results on three self-synthesized benchmarks, namely GenImage, AIGCDetectBenchmark, and DRCT-2M.
\item Section~\ref{sec:compare} presents the \textbf{computational overhead analysis}: the training and inference costs of AlignGemini with other VLM-based detectors, highlighting the benefits of our simplified training corpus and lightweight system prompt.
\item Section~\ref{sec:supp_task_model} provides \textbf{motivation analysis} of AlignGemini: extends the study to more backbones, validating the effectiveness of orthogonal supervision disentanglement.
\item Section~\ref{sec:aigi-now} describes the construction of the proposed AIGI-Now benchmark: the data sources, generation protocol, and semantic/pixel-discriminative subsets.
\item Section~\ref{sec:implementation} provides \textbf{implementation details of AlignGemini}, including the data construction process, the construction of DPO preference pairs, and the training and inference settings.
\item Section~\ref{sec:visualize} presents \textbf{qualitative visualizations} of AlignGemini's outputs: it shows representative predictions and response examples.

\end{itemize}


\section{Detailed Results of the Proposed AlignGemini}
\label{sec:detailed_results}

\paragraph{Per-Benchmark Comparison Results.}
We supplemented the detailed results of the individual subsets within each benchmark.
These detailed results are reported in Tables~\ref{tab:compare-genimage}--\ref{tab:compare-aigcdetectbenchmark}. 
GenImage and DRCT-2M are dominated by diffusion-based images, while AIGCDetectBenchmark primarily consists of GAN-generated images.
Comparisons of balanced accuracy on GenImage dataset are reported in Table~\ref{tab:compare-genimage}. Following BFree~\cite{guillaro2024bias}, in order to align images' format, we conduct JPEG-compression on synthetic images. Most methods suffer a significant decline when evaluated on these aligned images. Particularly, DRCT~\cite{chen2024drct} exhibits competitive performance on specific subsets such as Midjourney, SDv1.4, SDv1.5 and Wukong, but its performance degrades sharply on generators including ADM, GLIDE, and BigGAN.
In contrast, our AlignGemini demonstrate better generalization across diverse generators, achieving an overall balanced accuracy of 91.7\%. 
Comparisons of balanced accuracy on DRCT-2M dataset are reported in Table~\ref{tab:compare-drct}. AlignGemini achieves strong and stable performance across all DRCT-2M subsets, indicating robust generalization to a wide of diffusion generators. Specifically, our method consistently attains high balanced accuracy on both standard diffusion families and more challenging variants involving refinement, acceleration, and control mechanisms. Overall, AlignGemini reaches an average balanced accuracy of 98.1\%.
Across all benchmarks, AlignGemini achieves consistently strong performance and remains competitive with state-of-the-art methods.
Notably, AlignGemini attains competitive performance on AIGCDetectBenchmark, despite not using any GAN-generated images during training. 
This result indicates strong cross-architecture generalization beyond the diffusion models seen during training.

\begin{table}[h]
\centering
\begin{minipage}{\linewidth}
\centering
\caption{Comparison on GenImage. Following BFree~\cite{guillaro2024bias}, synthetic images are JPEG-compressed to match real images' format.}\vspace{-5pt}
\label{tab:compare-genimage}

\begin{adjustbox}{width=\linewidth}
\begin{tabular}{lccccccccc}
\toprule
Method & Midjourney & SDv1.4 & SDv1.5 & ADM & GLIDE & Wukong & VQDM & BigGAN & Avg. \\
\midrule
NPR \textsubscript{\textcolor{blue}{(CVPR'24)}} \cite{tan2024rethinking} 
& 53.4 & 55.1 & 55.0 & 43.8 & 41.2 & 57.4 & 48.4 & 57.7 & 51.5 $\pm$ 6.3 \\
UnivFD \textsubscript{\textcolor{blue}{(CVPR'23)}} \cite{ojha2023towards} 
& 55.1 & 55.6 & 55.7 & 62.5 & 61.3 & 61.1 & 76.9 & 84.4 & 64.1 $\pm$ 10.8 \\  
FatFormer \textsubscript{\textcolor{blue}{(CVPR'24)}} \cite{liu2024forgery} 
& 52.1 & 53.6 & 53.8 & 61.4 & 65.5 & 60.9 & 72.5 & 82.2 & 62.8 $\pm$ 10.4 \\
SAFE \textsubscript{\textcolor{blue}{(KDD'25)}} \cite{li2024improving} 
& 49.0 & 49.7 & 49.8 & 49.5 & 53.0 & 50.3 & 50.2 & 50.9 & 50.3 $\pm$ 1.2 \\
C2P-CLIP \textsubscript{\textcolor{blue}{(AAAI'25)}} \cite{tan2024c2p} 
& 56.6 & 77.5 & 76.9 & 71.6 & 73.5 & 79.4 & 73.7 & 85.9 & 74.4 $\pm$ 8.4 \\
AIDE \textsubscript{\textcolor{blue}{(ICLR'25)}} \cite{yan2024sanity} 
& 58.2 & 77.2 & 77.4 & 50.4 & 54.6 & 70.5 & 50.8 & 50.6 & 61.2 $\pm$ 11.9 \\
DRCT \textsubscript{\textcolor{blue}{(ICML'24)}} \cite{chen2024drct} 
& 82.4 & 88.3 & 88.2 & 76.9 & 86.1 & 87.9 & 85.4 & 87.0 & 84.7 $\pm$ 2.7 \\
AlignedForensics \textsubscript{\textcolor{blue}{(ICLR'25)}} \cite{rajan2025aligned} 
& \textbf{97.5} & \textbf{99.7} & \textbf{99.6} & 52.4 & 57.6 & \textbf{99.6} & 75.0 & 50.6 & 79.0 $\pm$ 22.7 \\
CO-SPY \textsubscript{\textcolor{blue}{(CVPR'25)}} \cite{cheng2025co}
& 58.5 & 63.1 & 69.8 & 72.5 & 81.6 & 92.5 & 82.2 & 90.3 & 76.3 $\pm$ 11.4 \\
BFree \textsubscript{\textcolor{blue}{(CVPR'25)}} \cite{guillaro2024bias} 
& 79.3 & 70.4 & 93.4 & \textbf{89.7} & \underline{86.8} & \underline{98.8} & \textbf{98.8} & \textbf{98.8} & \underline{89.5} $\pm$ 9.3 \\
\midrule
\rowcolor{myblue}
\textbf{AlignGemini}
& \underline{89.3} & \underline{89.6} & \underline{95.6} & \underline{77.2} & \textbf{90.3} & 98.3 & \underline{98.1} & \underline{98.4} & \textbf{91.7} $\pm$ 7.5 (\plus{+2.2})\\
\bottomrule 
\end{tabular}
\end{adjustbox}

\vspace{-10pt}
\end{minipage}

\end{table}

\begin{table*}[h]
  \centering
  \caption{Comparison of balanced accuracy on DRCT-2M.} 
  \vspace{-5pt}
  \label{tab:compare-drct}
  \begin{adjustbox}{width=0.9\linewidth}
  \begin{tabular}{lccccccccccccccccc}
  \toprule
  Method & LDM & SDv1.4 & SDv1.5 & SDv2 & SDXL & \makecell{SDXL-\\Refiner} & \makecell{SD-\\Turbo} & \makecell{SDXL-\\Turbo} & \makecell{LCM-\\SDv1.5} & \makecell{LCM-\\SDXL} & \makecell{SDv1-\\Ctrl} & \makecell{SDv2-\\Ctrl} & \makecell{SDXL-\\Ctrl} & \makecell{SDv1-\\DR} & \makecell{SDv2-\\DR} & \makecell{SDXL-\\DR} & Avg. \\
  \midrule
NPR \textsubscript{\textcolor{blue}{(CVPR'24)}} \cite{tan2024rethinking} 
& 33.0 & 29.1 & 29.0 & 35.1 & 33.2 & 28.4 & 27.9 & 27.9 & 29.4 & 30.2 & 28.4 & 28.3 & 34.7 & 67.9 & 67.4 & 66.1 & 37.3 $\pm$ 15.0 \\

UnivFD \textsubscript{\textcolor{blue}{(CVPR'23)}} \cite{ojha2023towards} 
& 85.4 & 56.8 & 56.4 & 58.2 & 63.2 & 55.0 & 56.5 & 53.0 & 54.5 & 65.9 & 68.0 & 65.4 & 75.9 & 64.6 & 56.2 & 53.9 & 61.8 $\pm$ 8.9 \\

FatFormer \textsubscript{\textcolor{blue}{(CVPR'24)}} \cite{liu2024forgery} 
& 55.9 & 48.2 & 48.2 & 48.2 & 48.2 & 48.3 & 48.2 & 48.2 & 48.3 & 50.6 & 49.7 & 49.9 & 59.8 & 66.3 & 60.6 & 56.0 & 52.2 $\pm$ 5.7 \\

SAFE \textsubscript{\textcolor{blue}{(KDD'25)}} \cite{li2024improving} 
& 50.3 & 50.1 & 50.0 & 50.0 & 49.9 & 50.1 & 50.0 & 50.0 & 50.1 & 50.0 & 49.9 & 50.0 & 54.7 & {98.2} & 98.5 & \underline{97.3} & 59.3 $\pm$ 19.2 \\

C2P-CLIP \textsubscript{\textcolor{blue}{(AAAI'25)}} \cite{tan2024c2p} 
& 83.0 & 51.7 & 51.7 & 52.9 & 51.9 & 64.6 & 51.7 & 50.6 & 52.0 & 66.1 & 56.9 & 54.7 & 77.8 & 67.2 & 57.1 & 56.7 & 59.2 $\pm$ 9.9 \\

AIDE \textsubscript{\textcolor{blue}{(ICLR'25)}} \cite{yan2024sanity} 
& 64.4 & 74.9 & 75.1 & 58.5 & 53.5 & 66.3 & 52.8 & 52.8 & 70.0 & 54.3 & 65.9 & 53.6 & 53.9 & 95.3 & 73.3 & 69.0 & 64.6 $\pm$ 11.8 \\

DRCT \textsubscript{\textcolor{blue}{(ICML'24)}} \cite{chen2024drct} 
& 96.7 & 96.3 & 96.3 & 94.9 & 96.2 & 93.5 & 93.4 & 92.9 & 91.2 & 95.0 & 95.6 & 92.7 & 92.0 & 94.1 & 69.6 & 57.4 & 90.5 $\pm$ 7.4 \\

AlignedForensics \textsubscript{\textcolor{blue}{(ICLR'25)}} \cite{rajan2025aligned} 
& \textbf{99.9} & \textbf{99.9} & \textbf{99.9} & \textbf{99.6} & 90.2 & 81.3 & \textbf{99.7} & 89.4 & \textbf{99.7} & 90.0 & \textbf{99.9} & 99.2 & 87.6 & \textbf{99.9} & \underline{99.8} & 92.6 & 95.5 $\pm$ 6.1 \\

CO-SPY \textsubscript{\textcolor{blue}{(CVPR'25)}} \cite{cheng2025co} 
& 72.2 & \underline{99.4} & 82.4 & 99.3 & 73.7 & 96.7 & 70.9 & \underline{97.2} & 94.8 & 51.0 & 51.5 & \underline{99.4} & \textbf{99.4} & 50.7 & 83.9 & 91.1 & 83.1 $\pm$ 16.7 \\

BFree \textsubscript{\textcolor{blue}{(CVPR'25)}} \cite{guillaro2024bias}
& \underline{99.5} & 99.3 & \underline{99.3} & \underline{99.4} & \textbf{99.5} & \textbf{99.0} & \underline{99.5} & \textbf{98.0} & \underline{99.4} & \textbf{99.4} & \underline{99.4} & \textbf{99.5} & \textbf{99.4} & 96.4 & \textbf{99.9} & \textbf{99.2} & \textbf{99.1} $\pm$ 0.8 \\
  \midrule



  \rowcolor{myblue}
Ours 
& 99.1 & 98.7 & 98.8 & 98.2 & \underline{98.0} 
& \underline{98.9} & 97.7 & 94.6 & 96.4 & \underline{98.3}
& 98.6 & 99.0 & \underline{99.2} & \underline{98.9} & 99.2 
& {96.2} & \underline{98.1} $\pm$ 1.3 (\minus{-1.0})
\\
\bottomrule
  \end{tabular}
  \end{adjustbox}
   \vspace{-5pt}
\end{table*}

\begin{table*}[h]
\centering
\caption{Comparison of balanced accuracy on AIGCDetectBenchmark.}
\vspace{-5pt}
\label{tab:compare-aigcdetectbenchmark}
\begin{adjustbox}{width=0.9\linewidth}
\begin{tabular}{lcccccccccccccccccl}
\toprule
Method & ADM & DALLE2 & GLIDE & Midjourney & VQDM & BigGAN & CycleGAN & GauGAN & ProGAN & SDXL & SD14 & SD15 & StarGAN & StyleGAN & StyleGAN2 & WFR & Wukong & Avg. \\
\midrule
NPR \textsubscript{\textcolor{blue}{(CVPR'24)}} \cite{tan2024rethinking}
& 43.8 & 20.0 & 41.2 & 53.4 & 48.4 & 53.1 & 76.6 & 42.2 & 58.7 & 59.6 & 55.1 & 55.0 & 67.4 & 57.9 & 54.6 & 58.8 & 57.4 & 53.1 $\pm$ 12.2 \\

UnivFD \textsubscript{\textcolor{blue}{(CVPR'23)}} \cite{ojha2023towards}
& 62.5 & 50.0 & 61.3 & 55.1 & 76.9 & 87.5 & \underline{96.9} & \underline{98.8} & \textbf{99.4} & 58.2 & 55.6 & 55.7 & 95.1 & 80.0 & 69.4 & 69.2 & 61.1 & 72.5 $\pm$ 17.3 \\

FatFormer \textsubscript{\textcolor{blue}{(CVPR'24)}} \cite{liu2024forgery}
& \underline{80.2} & 68.5 & \textbf{91.1} & 54.4 & 88.0 & \textbf{99.2} & \textbf{99.5} & \textbf{99.1} & 98.5 & 71.7 & 67.5 & 67.2 & \underline{99.4} & \textbf{98.0} & \textbf{98.8} & \underline{88.3} & 75.6 & 85.0 $\pm$ 14.9 \\

SAFE \textsubscript{\textcolor{blue}{(KDD'25)}} \cite{li2024improving}
& 49.5 & 49.5 & 53.0 & 49.0 & 50.2 & 52.2 & 51.9 & 50.0 & 50.0 & 49.8 & 49.7 & 49.8 & 50.1 & 50.0 & 50.0 & 49.8 & 50.3 & 50.3 $\pm$ 1.1 \\

C2P-CLIP \textsubscript{\textcolor{blue}{(AAAI'25)}} \cite{tan2024c2p}
& 71.6 & 52.3 & 73.5 & 56.6 & 73.7 & \underline{98.4} & 96.8 & \underline{98.8} & \underline{99.3} & 62.3 & 77.5 & 76.9 & \textbf{99.6} & \underline{93.1} & 79.4 & \textbf{94.8} & 79.4 & 81.4 $\pm$ 15.6 \\

AIDE \textsubscript{\textcolor{blue}{(ICLR'25)}} \cite{yan2024sanity}
& 52.9 & 51.1 & 60.2 & 49.8 & 69.3 & 70.1 & 93.6 & 60.6 & 89.0 & 49.6 & 51.6 & 51.0 & 72.1 & 66.5 & 59.0 & 80.6 & 54.5 & 63.6 $\pm$ 13.9 \\

DRCT \textsubscript{\textcolor{blue}{(ICML'24)}} \cite{chen2024drct}
& 79.9 & \underline{89.2} & 89.2 & 85.5 & \underline{88.6} & 81.4 & 91.0 & 93.8 & 71.1 & 88.3 & 91.4 & 91.0 & 53.0 & 62.7 & 63.8 & 73.9 & 90.8 & 81.4 $\pm$ 12.2 \\

AlignedForensics \textsubscript{\textcolor{blue}{(ICLR'25)}} \cite{rajan2025aligned}
& 51.6 & 52.0 & 55.6 & \textbf{96.2} & 72.1 & 51.2 & 49.5 & 50.8 & 50.7 & 95.1 & \textbf{99.7} & \textbf{99.6} & 53.8 & 52.7 & 51.6 & 50.0 & \textbf{99.6} & 66.6 $\pm$ 21.6 \\

CO-SPY \textsubscript{\textcolor{blue}{(CVPR'25)}} \cite{cheng2025co}
& 58.5 & 84.9 & 81.7 & 69.8 & 72.6 & 70.5 & 53.5 & 68.1 & 74.0 & 81.4 & 92.5 & 92.2 & 62.8 & 60.0 & 60.1 & 60.9 & 90.2 & 72.5 $\pm$ 12.6 \\

BFree \textsubscript{\textcolor{blue}{(CVPR'25)}} \cite{guillaro2024bias}
& 79.3 & 85.8 & 86.8 & 93.4 & \textbf{89.7} & 91.9 & 76.8 & 97.0 & 95.9 & \textbf{99.3} & \underline{98.8} & \underline{98.8} & 86.1 & 82.1 & 78.0 & 60.5 & \underline{98.8} & \textbf{88.2} $\pm$ 10.5 \\
\midrule



\rowcolor{myblue}
\textbf{AlignGemini}
& \textbf{89.3} & \textbf{94.9} & \underline{90.3} & \underline{95.6} & 77.2 & 90.4 & 74.4 & 92.5 & 91.0 & \underline{99.1} & 98.3 & 98.1 & 72.9 & 86.6 & \underline{89.2} & 52.6 & 98.4 & \underline{87.7} $\pm$ 12.2(\minus{-0.5}) \\
\bottomrule
\end{tabular}
\end{adjustbox}
  \vspace{-5pt}
\end{table*}

\section{Overhead Comparison}
\label{sec:compare}
\paragraph{Indirect Comparison of Overall Pipeline Overhead.}
A practical consideration for VLM-based detectors is the end-to-end pipeline overhead, including data construction, training, and inference. Table~\ref{tab:overhead} reports indirect statistics for VLM-based detectors; because some official implementations are not publicly available, we compare them qualitatively rather than via unified wall-clock measurements.\footnote{The average tokens of IVY-FAKE and FakeVLM are obtained by referring to the IVY-FAKE~\cite{zhang2025ivy} paper.} The results highlight three advantages of AlignGemini-VLM: (i) its VLM branch is trained with fewer samples than competing VLM-based detectors, reducing training overhead; (ii) it does not require manual annotation checking, since semantically implausible images can be reliably labeled by the model itself; and (iii) it adopts a concise system prompt that forces the VLM to focus exclusively on identifying semantic flaws, while delegating the pixel-artifact detection task—which is particularly difficult for VLMs and prone to hallucinations—to a dedicated expert model. As a result, the VLM branch exploits its inherent semantic perception capabilities to detect high-level semantic inconsistencies "at a glance". Moreover, the overhead of AlignGemini’s expert branch, which mainly consists of VAE reconstruction and LoRA fine-tuning of DINOv2, is negligible compared with that of the VLM branch.

\begin{table*}[h]
\centering
\caption{Comparison of training data and inference cost for VLM-based detectors.}
\label{tab:overhead}
\begin{adjustbox}{width=0.98\linewidth}
\begin{tabular}{llcccccc}
\toprule
 & & \multicolumn{3}{c}{Training} & \multicolumn{3}{c}{Inference} \\
\cmidrule(lr){3-5} \cmidrule(lr){6-8}
Method & Backbone & Training samples & Annotation type & Annotation check & Avg. tokens $\downarrow$ & Sec. / token & Sec. / Ans. $\downarrow$ \\
\midrule
IVY-FAKE~\cite{zhang2025ivy}      & Ivy-VL-LLaVA         & 150K & Automated by Gemini 2.5 pro & LLM & 530 & 0.0424 & 22.47 \\ 
FakeVLM~\cite{SpotFake}  & LLaVA-v1.5-7B & 95K & Automated by 3 MLLMs & LLM & 120 & 0.0128 & 1.54 \\
AIGI-Holmes~\cite{li2025aigi}    & LLaVA-v1.6-Mistral-7B & 69K & Automated by 4 MLLMs & LLM + Human & 181 & \textbf{0.0057} &  1.03  \\ 
Forensic-Chat~\cite{lin2025seeing} & Qwen2.5-VL-7B & 61K & Automated by Gemini 2.5 pro & None & 307 & 0.0062 & 1.90 \\
\midrule
\rowcolor{myblue}
\textbf{AlignGemini-VLM}      & Qwen2.5-VL-7B        & \textbf{10K} & Automated by Qwen2.5-VL-7B & None & \textbf{97} & 0.0062 & \textbf{0.60} \\ 
\bottomrule
\end{tabular}
\end{adjustbox}
\end{table*}

\begin{figure*}[h]
\centering
\includegraphics[width=0.95\linewidth]{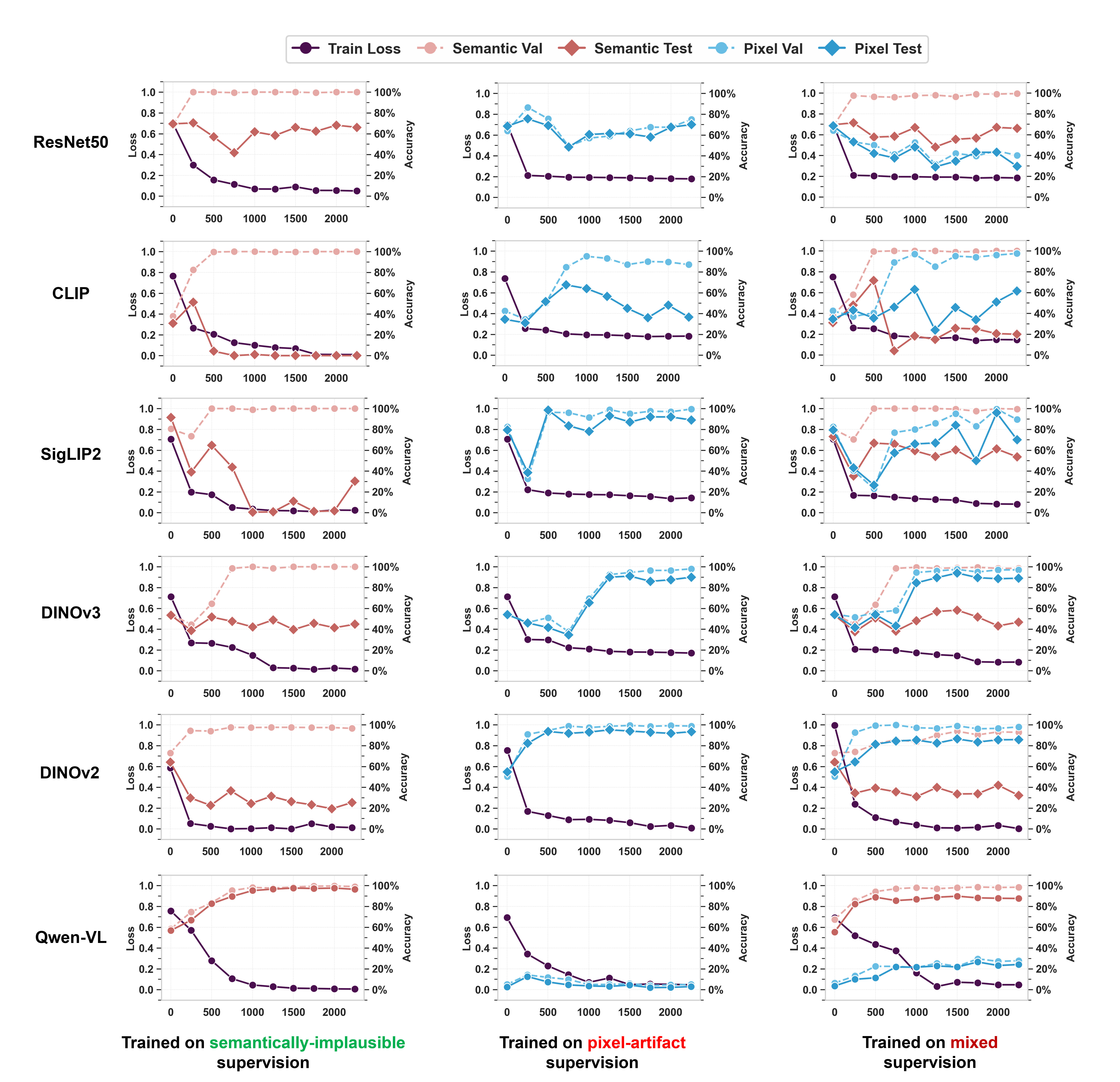}
\caption{Controlled studies on the effect of supervision structure on six models.}
\label{fig:extented}
\end{figure*}

\section{Extended Study on Orthogonal Supervision Disentanglement}
\label{sec:supp_task_model}

Figure~3 of the main paper presents detection performance of DINOv2 and VLM(Qwen-VL) under each supervision regime.
This section extends the analysis to four additional models: CLIP~\cite{CLIP}, SigLIP~\cite{tschannen2025siglip}, DINOv3~\cite{oquab2023dinov2}, and ResNet-50~\cite{he2016deep}, to show their performance under the same training strategies.

\paragraph{Experimental Setup}
We evaluate on six models:
ResNet-50~\cite{he2016deep},
CLIP~\cite{CLIP}, SigLIP~\cite{tschannen2025siglip},
DINOv2~\cite{oquab2023dinov2}, DINOv3,
and Qwen-VL~\cite{qwen25}.
Each model is trained under three supervision regimes following Section~4:
(i)~a \emph{semantically-implausible set}, in which images contain semantic inconsistencies produced by GPT-4o and are post-processed to reduce low-level cues;
(ii)~a \emph{pixel-artifact set}, in which images are generated via VAE reconstruction to preserve semantics while modifying pixel statistics;
and (iii)~a \emph{mixed set} combines both.

\paragraph{Results and Analysis.}
The training curves for all experiments are shown in Figure~\ref{fig:extented}, and a qualitative summary is provided in Table~\ref{tab:capability_summary_text}.
\noindent\textbf{ResNet-50.}
The validation and test curves remain relatively close in several settings, suggesting that it captures part of the signal in both cases.
However, its overall performance does not indicate strong generalization on either subtask.
\noindent\textbf{CLIP and SigLIP.}
Under semantically-implausible supervision, both models fit the training data quickly but show limited generalization to the unseen generator.
This suggests that the semantic representations learned by encoders in this setting are less suitable for robust cross-generator semantically-implausible detection.
In contrast, the VLM remains consistently stronger on the semantic subtask and shows better generalization behavior.
\noindent\textbf{DINOv2 and DINOv3.}
Both of them remain relatively stable under pixel-artifact supervision and maintain strong performance on the pixel-artifact detection subtask.
Among them, DINOv2 shows more reliable behavior, with a smaller gap between validation and test, which is consistent with its stronger capability for pixel-artifact detection.
Under mixed supervision, no model shows clear advantage on both subtasks at the same time.
Instead, the mixed setting tends to weaken the stronger capability of each model rather than improve both.
Overall, the trends in Figure~\ref{fig:extented} and Table~\ref{tab:capability_summary_text} are consistent with the orthogonal supervision strategy.
The VLM is more suitable for semantically-implausible detection, while conventional vision models (e.g., DINOv2) are more suitable for pixel-artifact detection.
The proposed orthogonal supervision disentanglement provides a favorable solution for generalizable AIGI detection.

\begin{table}[h]
\centering
\caption{\textbf{Qualitative summary of encoder capability under orthogonal supervision.}
The capability levels are summarized according to cross-generator semantically-implausible detection and pixel-artifact detection performance.}
\label{tab:capability_summary_text}
\setlength{\tabcolsep}{6pt}
\begin{adjustbox}{width=\linewidth}
\begin{tabular}{lcccccc}
\toprule
\textbf{Capability} & \textbf{DINOv3} & \textbf{DINOv2} & \textbf{ResNet-50} & \textbf{CLIP} & \textbf{SigLIP2} & \textbf{Qwen-VL} \\
\midrule
Semantically-implausible detection
& \Moderate
& \Moderate
& \Weak
& \Weak
& \Moderate
& \Strong \\
Pixel-artifact detection
& \Strong
& \Strong
& \Weak
& \Moderate
& \Moderate
& \Weak \\
\bottomrule
\end{tabular}
\end{adjustbox}
\end{table}

\begin{figure*}[t!]
    \centering
    \includegraphics[width=1.0\linewidth]{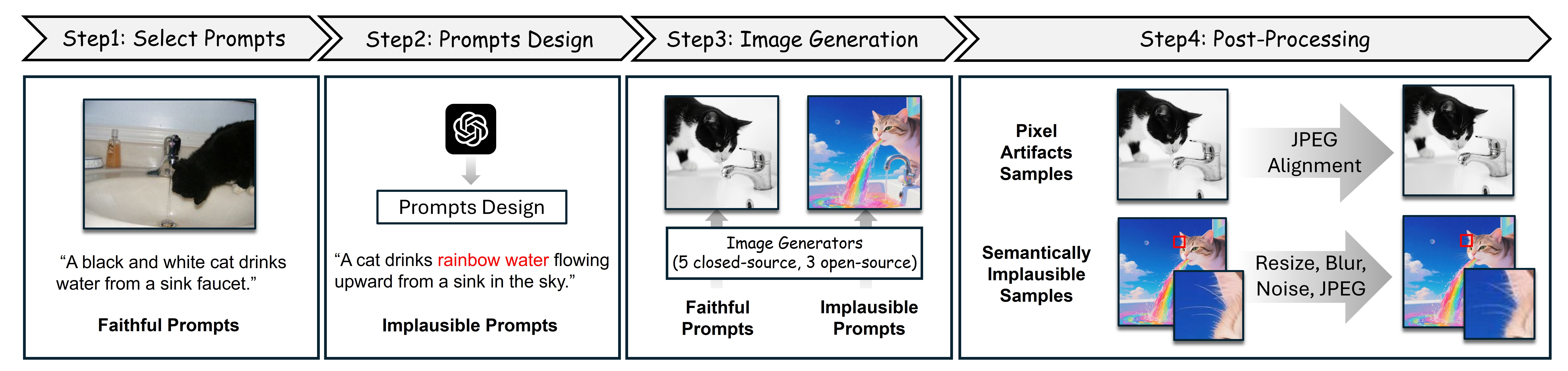}
    \caption{Overall pipeline for constructing the AIGI-Now benchmark.}
    \label{fig:pipeline_aigi-now}
\end{figure*}

\section{Design of the AIGI-Now Benchmark}
\label{sec:aigi-now}

\paragraph{Overview of Employed Generative Models.}
AIGI-Now covers widely used modern generators, including eight generative models:
\begin{itemize}
    \item Nano Banana\footnote{https://deepmind.google/models/gemini/image/} (Google Gemini 2.5 Flash Image): a state-of-the-art commercial image generation and editing model from Google Gemini.
    \item Jimeng\footnote{https://jimeng.jianying.com/} (jimeng-t2i-v30): a multimodal commercial model from ByteDance supporting high-quality text-to-image generation.
    \item Kling\footnote{https://app.klingai.com/} (kling-v2): a commercial model developed by Kuaishou for high-quality image and video generation.
    \item Minimax\footnote{https://minimaxi.com/} (image-01): MiniMax's flagship text-to-image model optimized for realistic, prompt-faithful imagery.
    \item Flux Pro\footnote{https://bfl.ai/models/flux-pro} (flux1.1-pro): a text-to-image model from Black Forest Labs combining high visual quality with fast inference.
    \item Flux Dev\footnote{https://huggingface.co/black-forest-labs/FLUX.1-dev}: a general-purpose FLUX.1 model for high-quality text-to-image generation.
    \item Flux Krea\footnote{https://huggingface.co/black-forest-labs/FLUX.1-Krea-dev}: a FLUX.1 variant tailored for creative and stylized image generation.
    \item Flux Kontext\footnote{https://huggingface.co/black-forest-labs/FLUX.1-Kontext-dev}: a FLUX.1 model for image generation and editing that can preserve contextual details and text–image consistency.
\end{itemize}

\paragraph{Overall Pipeline of Constructing AIGI-Now.}
Figure~\ref{fig:pipeline_aigi-now} illustrates the overall pipeline for constructing AIGI-Now. For each generative model, the construction proceeds in four stages: (i) we randomly sample real images from the COCO test split; (ii) starting from the associated captions, we construct two types of prompts, namely semantically faithful prompts (for pixel-artifact subsets) and semantically implausible prompts (for semantic subsets), as described above; (iii) we employ generative model to synthesize images conditioned on these prompts; and (iv) we form pixel-artifact and semantic subsets with targeted pre- and post-processing. For the pixel-artifact subsets, which consist of real images and semantically faithful synthetic images, we align the formats of synthetic samples with their real counterparts by applying the same JPEG compression settings, thereby mitigating format bias. For the semantic subsets, which consist of real images and semantically implausible synthetic images, we apply a suite of heavy post-processing operations to both real and synthetic images to prevent detectors from exploiting low-level artifacts. Specifically, each image undergoes the following operations: (i) it is first downsampled to 40\% of its original resolution; (ii) Gaussian noise with a standard deviation of 0.1 is injected; (iii) a Gaussian blur with a radius of 0.20 is applied; (iv) the image is re-encoded using JPEG compression with a quality factor of 80; and (v) the degraded image is upsampled back to its original resolution. These transformations substantially distort low-level statistics while preserving high-level semantics, forcing detectors to rely on robust forensic cues rather than brittle artifacts.

\paragraph{Prompt Design.}
AIGI-Now uses the widely adopted COCO dataset as the source of real images. Starting from COCO captions, we design two types of prompts. (i) \emph{Semantically faithful} prompts, used to construct pixel-artifact subsets: we directly adopt the original COCO captions to synthesize images whose semantic content roughly matches the corresponding real images. (ii) \emph{Semantically implausible} prompts, used to construct semantic subsets: we leverage LLM to inject physically impossible, surreal, or logically contradictory attributes into the original COCO captions while preserving the main entities. Representative examples of semantically implausible prompts are shown in the prompt box in Figure~\ref{promptbox:implausible_prompts}.

\newtcolorbox{promptbox}{
  colback=gray!5,
  colframe=gray!5, 
  boxrule=0pt,      
  arc=10pt,         
  boxsep=5pt,       
  left=10pt,        
  right=10pt,       
  top=10pt,         
  bottom=10pt,      
  fontupper=\footnotesize
}

\begin{figure}
\begin{promptbox}
\centering

\textbf{Semantically Implausible Prompts}

\begin{enumerate}[leftmargin=*, label=\textbf{\arabic*.}, itemsep=2pt, topsep=2pt]
\item A woman sitting at a table in front of a pile of \emph{glowing} luggage.
\item A pile of \emph{melting} luggage in a room with two women nearby.
\item The man stands holding a surfboard as he looks out at a \emph{sea of stars}.
\item A couple celebrating a birthday in a kitchen \emph{underwater}.
\item Lots of skiers standing atop a hill of \emph{giant crystals}.
\item Skiers and snowboarders standing on a mountain of \emph{giant clocks}.
\item a young boy standing in front of a ball \emph{made of stars}.
\item A man with fruit \emph{made of glass} on the back of a pickup truck.
\item A group of skiers gather on a hill \emph{made of clouds}.
\item A man holding a \emph{clock-shaped} object in his hand.
\item A man playing tennis on \emph{clouds}.
\item A living room with a \emph{floating} TV and a \emph{glowing} wheel.
\item A bathroom overlooking an \emph{upside-down} ocean.
\item A city intersection where traffic lights \emph{rain upward}.
\item A horse in a snowy pasture with a \emph{neon-lit} fence.
\item Man at market with \emph{glowing blue} bananas.
\item A boy jumping on a skateboard with \emph{wings}.
\item A farmer's market filled with \emph{giant} bananas and \emph{shadowless} people.
\item A group beside a \emph{glowing} bench with \emph{transparent} frisbees.
\item Breakfast and juice served \emph{on a cloud}.
\item A bench on a beach \emph{on the moon}.
\item An empty street with \emph{floating} buildings and \emph{upside-down} cars.
\item A black and white image of an old person \emph{melting into shadows}.
\item A city intersection sign made of \emph{ice} in front of a building of \emph{clouds}.
\item A bus stopped on the side of a road that \emph{melts into a river}.
\item A white dog made of \emph{clouds} standing on a \emph{floating} wooden bench.
\item Stop sign with a tree branch made of \emph{liquid metal} touching it.
\item A man sitting on a bench with an ocean of \emph{liquid glass} behind him.
\item A giant clock on the very top of a building \emph{growing like a tree}.
\item A kitchen with a \emph{glowing} refrigerator and \emph{floating} counter chairs.

  ......
\end{enumerate}
\end{promptbox}
\caption{Illustrative prompts used in AIGI-Now.}
\label{promptbox:implausible_prompts}
\end{figure}

\begin{figure*}[h]
    \centering
    \includegraphics[width=0.9\linewidth]{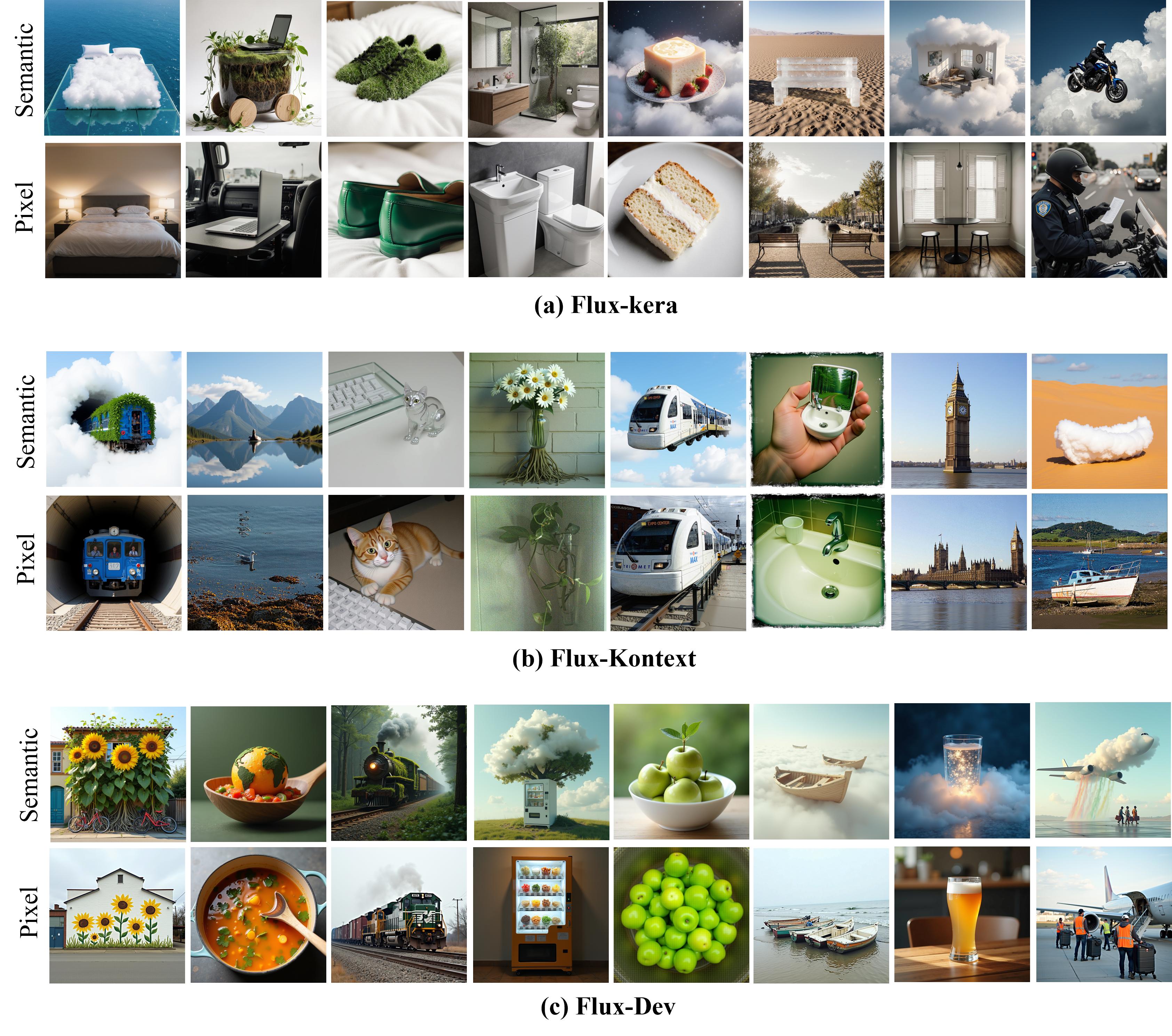}    
    \caption{Representative AIGI-Now examples synthesized by Flux-Kera, Flux-Kontext, and Flux-Dev, illustrating the high visual fidelity and stylistic diversity of recent Flux generators.}
    \label{fig:dataset_visualization_examples1}   
\end{figure*}

\begin{figure*}[h!]
    \centering
    \includegraphics[width=0.65\linewidth]{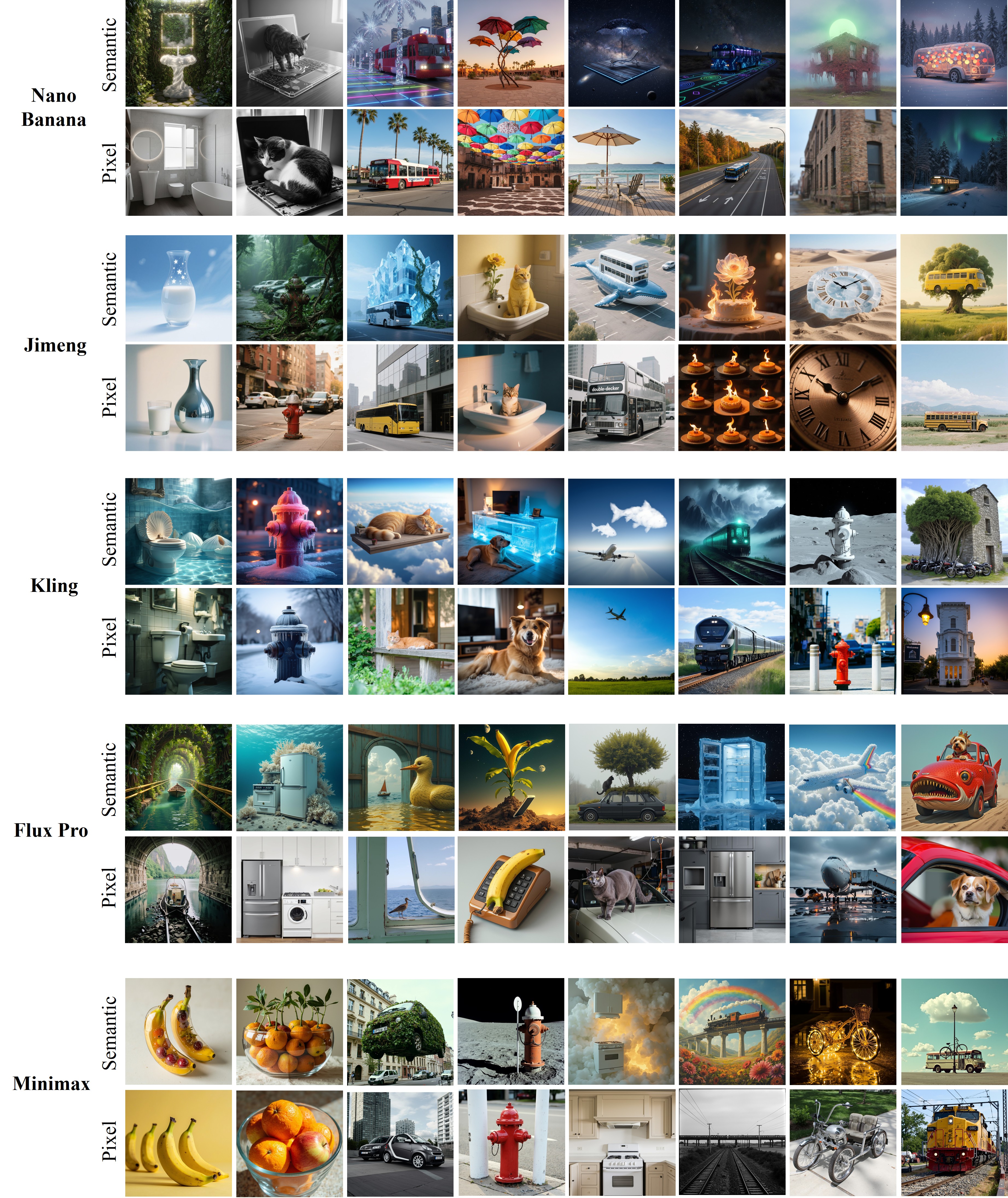}    
    \vspace{-10pt}
     \caption{Representative AIGI-Now examples generated by five recent closed-source models, showcasing diverse scenes and styles that increase the difficulty of robust AI-generated image detection.}
    \label{fig:dataset_visualization_examples2}   
\end{figure*}

\paragraph{Visualization of the AIGI-Now Benchmark.}
Figures~\ref{fig:dataset_visualization_examples1} and \ref{fig:dataset_visualization_examples2} illustrate the semantic and pixel-artifact subsets of AIGI-Now, spanning eight different generative models. These examples provide an intuitive view of high-quality synthetic images produced by recent generative models.

\section{Implementation Details}
\label{sec:implementation}
In this section, we present additional implementation details, including the construction of the training data, construction of DPO preference pairs, and the training and inference settings.

\paragraph{Image Source.}
\textbf{(i) Semantic supervision set:}
The training set for the VLM branch consists of 10,000 images. Specifically, we randomly sample 5,000 synthetic images from thFe \textit{Surreal Fantasy} subset of Echo-4o~\cite{ye2025echo} and 5,000 real images from Unsplash~\cite{unsplash_data_2025}.
\textbf{(ii) Pixel-artifact supervision set:}
The training set for the expert branch consists of 11.8k MSCOCO images and their synthetic counterparts generated using the Stable Diffusion 2.1 VAE.

\begin{figure*}[h]
  \centering

  \begin{subfigure}[t]{0.48\textwidth}
    \centering
    \begin{promptbox}
      Role: AI visual assistant.\\
      You are an AI visual assistant that analyzes a single image to determine whether it is authentic or AI-generated.\\
      \textbf{Tasks}
      \begin{enumerate}[leftmargin=15pt]
        \item \textless rationale\textgreater\ section:
        \begin{itemize}[leftmargin=12pt]
          \item Provide concise, evidence-based reasoning in fewer than 120 words.
          \item Focus strictly on semantic or logical reasoning.
        \end{itemize}

        \item \textless answer\textgreater\ section:
        \begin{itemize}[leftmargin=12pt]
          \item Only output ``This is an authentic image.''
        \end{itemize}
      \end{enumerate}
      \textbf{Output Structure (must match exactly):}\\
      \textless rationale\textgreater\textless/rationale\textgreater\\
      \textless answer\textgreater``This is an authentic image.''\textless/answer\textgreater
    \end{promptbox}
    \label{fig:dpo_prompt_real}
  \end{subfigure}
  \hfill
  \begin{subfigure}[t]{0.48\textwidth}
    \centering
    \begin{promptbox}
      Role: AI visual assistant.\\
      You are an AI visual assistant that analyzes a single image to determine whether it is authentic or AI-generated.\\
      \textbf{Tasks}
      \begin{enumerate}[leftmargin=15pt]
        \item \textless rationale\textgreater\ section:
        \begin{itemize}[leftmargin=12pt]
          \item Provide concise, evidence-based reasoning in fewer than 120 words.
          \item Focus strictly on semantic or logical reasoning.
        \end{itemize}

        \item \textless answer\textgreater\ section:
        \begin{itemize}[leftmargin=12pt]
          \item Only output ``This is an AI-generated image.''
        \end{itemize}
      \end{enumerate}
      \textbf{Output Structure (must match exactly):}\\
      \textless rationale\textgreater\textless/rationale\textgreater\\
      \textless answer\textgreater``This is an AI-generated image.''\textless/answer\textgreater
    \end{promptbox}
    \label{fig:dpo_prompt_fake}
  \end{subfigure}

  \caption{Prompt templates used to construct DPO preference pairs.}
  \label{fig:dpo_prompt_templates}
\end{figure*}

\paragraph{Construction of DPO Preference Pairs.}
To fine-tune the VLM branch with semantic supervision, we construct DPO preference pairs from the semantic supervision set. 
Specifically, we use Qwen2.5-VL-7B~\cite{qwen25} to generate responses under a unified prompt template. The model is instructed to analyze a single image, provide a concise rationale based strictly on semantic or logical evidence, and then output a fixed verdict. The prompt template is shared across all samples, and the only label-dependent modification is the assigned target verdict: for real images, the target answer is specified as \textit{``This is an authentic image.''}; for AI-generated images, it is specified as \textit{``This is an AI-generated image.''}
Then, we construct DPO preference pairs according to the ground-truth label of each image. The response whose verdict matches the image label is treated as the preferred response, while the opposite one is treated as the dispreferred response.

\paragraph{Training Setting.}
AlignGemini is trained with two branches. \textbf{VLM branch:} We adopt Qwen2.5-VL-7B as the semantic branch and fine-tune it with DPO on the semantic supervision set. Specifically, the training set consists of 5,000 randomly sampled real images from Unsplash and 5,000 randomly sampled synthetic images from Echo-4o. The DPO training is implemented based on the MS-SWIFT framework~\cite{zhao2025swift}. We use LoRA fine-tuning with rank 16 and $\alpha=32$, a learning rate of $1\times10^{-6}$, a batch size of 8, and a DPO coefficient $\beta=0.05$, and train the model for one epoch. \textbf{Expert branch:} We adopt DINOv2-ViT-L/14 with a classification head as the pixel-artifact expert and fine-tune it on a pure pixel-artifact corpus constructed from 11.8k COCO training images and their corresponding VAE-reconstructed counterparts. This branch is fine-tuned with LoRA, using rank 8 and $\alpha=1.0$, with a learning rate of $1\times10^{-4}$ and a batch size of 16.



\paragraph{Inference Setting.}
During inference, we apply logical OR fusion: an image is classified as real only if both branches predict real; otherwise, it is classified as AI-generated. For the VLM branch, the prediction is obtained from the fixed verdict in the \texttt{<answer>} field. The expert branch independently predicts whether pixel-level generation artifacts are present.

\paragraph{System Prompt of AlignGemini's VLM}
Figure~\ref{promptbox:system_prompts} illustrates the system prompt used for AlignGemini's VLM detector.
Unlike prior VLM detectors (e.g., AIGI-Holmes), whose system prompts exhaustively enumerate possible visual artifacts and require the model to check them one by one, our prompt explicitly instructs the VLM to (i) focus strictly on semantic and logical reasoning and (ii) provide concise, evidence-based rationales in fewer than 120 words.
This design steers the model toward detecting semantic inconsistencies rather than low-level pixel artifacts, reduces inference cost by generating substantially fewer tokens, and alleviates hallucinations by deliberately avoiding instructions to attend to fine-grained visual artifacts.


\begin{figure}[h]
  \centering
  \begin{promptbox}
    Role: AI visual assistant.\\
    You are an AI visual assistant that analyzes a single image to determine whether it is authentic or AI-generated.\\
    \textbf{Tasks}
    \begin{enumerate}[leftmargin=15pt]
      \item \textless rationale\textgreater\ section:
      \begin{itemize}[leftmargin=12pt]
        \item Provide concise, evidence-based reasoning in fewer than 120 words.
        \item Focus strictly on semantic or logical reasoning.
      \end{itemize}

      \item \textless answer\textgreater\ section:
      \begin{itemize}[leftmargin=12pt]
        \item If real: only output ``This is an authentic image.''
        \item If AI-generated: only output ``This is an AI-generated image.''
      \end{itemize}
    \end{enumerate}
    \textbf{Output Structure (must match exactly):}\\
    \textless rationale\textgreater\textless/rationale\textgreater\\
    \textless answer\textgreater\textless/answer\textgreater
  \end{promptbox}
  \caption{System prompt.}
  \label{promptbox:system_prompts}
\end{figure}


\section{Visualization of AlignGemini Outputs}
\label{sec:visualize}

\begin{figure*}[h]
\centering
\includegraphics[width=0.8\linewidth]{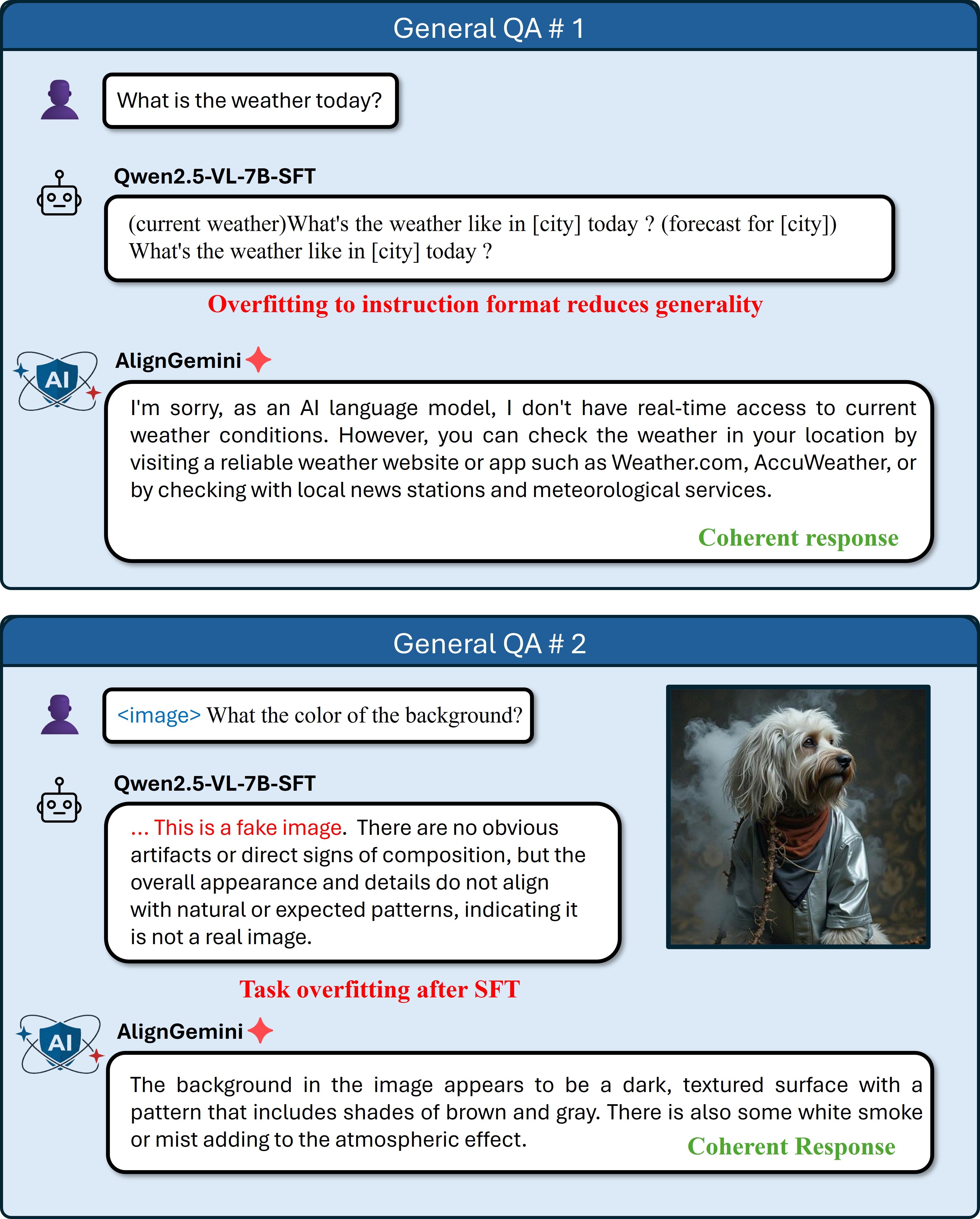}
\caption{Examples of general multimodal conversations. It indicates that the general capabilities of AlignGemini are largely preserved after adaptation for AIGI detection.}
\label{fig:general_capabilities}
\end{figure*}

\paragraph{AlignGemini preserves general capabilities of the VLM.}
Table~\ref{tab:general} quantitatively compares different VLM-based detectors in terms of their general multimodal understanding capabilities.
The results show that AlignGemini preserves substantially more of the VLM's general capabilities than other detectors, mainly due to two factors: (i) our training strategy that reinforces the VLM's inherent strengths and (ii) preference-guided DPO fine-tuning.
Specifically, AlignGemini does not force the VLM to learn misaligned skills such as explicit pixel-artifact identification, and it avoids supervised fine-tuning on curated "hard" samples that typically damage general multimodal understanding.
Figure~\ref{fig:general_capabilities} provides qualitative visual examples.
These examples show that, despite being adapted for AIGI detection, the VLM still preserves strong general multimodal capabilities, because our training strategy maintains its inherent strengths rather than pushing it to learn tasks misaligned with those strengths.

\begin{table}[h]
\centering
\scriptsize
\caption{Comparison of AlignGemini and baseline VLM-based detectors on general multimodal understanding benchmarks. Results are referred from~\cite{lin2025seeing}.}
\begin{tabular}{l|cccc}
\toprule
Method & BLINK & RealWorldVQA & MME & MMT-Bench\textsubscript{VAL} \\
\midrule
\rowcolor{gray!10}
\multicolumn{5}{l}{\textbf{General MLLM}} \\
\addlinespace[3pt]
Qwen-2.5-VL-3B & 0.4750 & 0.6588 & 1590 & 0.6025 \\
Qwen-2.5-VL-7B & 0.5481 & 0.6758 & 1677 & 0.5948 \\
LLaVA-1.5-7B   & 0.4171 & 0.5424 & 1436 & 0.4713 \\
\midrule
\rowcolor{gray!10}
\multicolumn{5}{l}{\textbf{MLLM-based AIGI detector}} \\
\addlinespace[3pt]
FakeVLM        & 0.3761 & 0.5385 & 1221 & 0.4445 \\
Forensic-Chat  & 0.5139 & 0.6745 & 1625 & 0.5849 \\
\rowcolor{myblue}
AlignGemini    & \textbf{0.5321} & \textbf{0.6759} & \textbf{1643} & \textbf{0.5911} \\
\bottomrule
\end{tabular}
\label{tab:general}
\end{table}

\begin{figure*}
\centering
\includegraphics[width=0.8\linewidth]{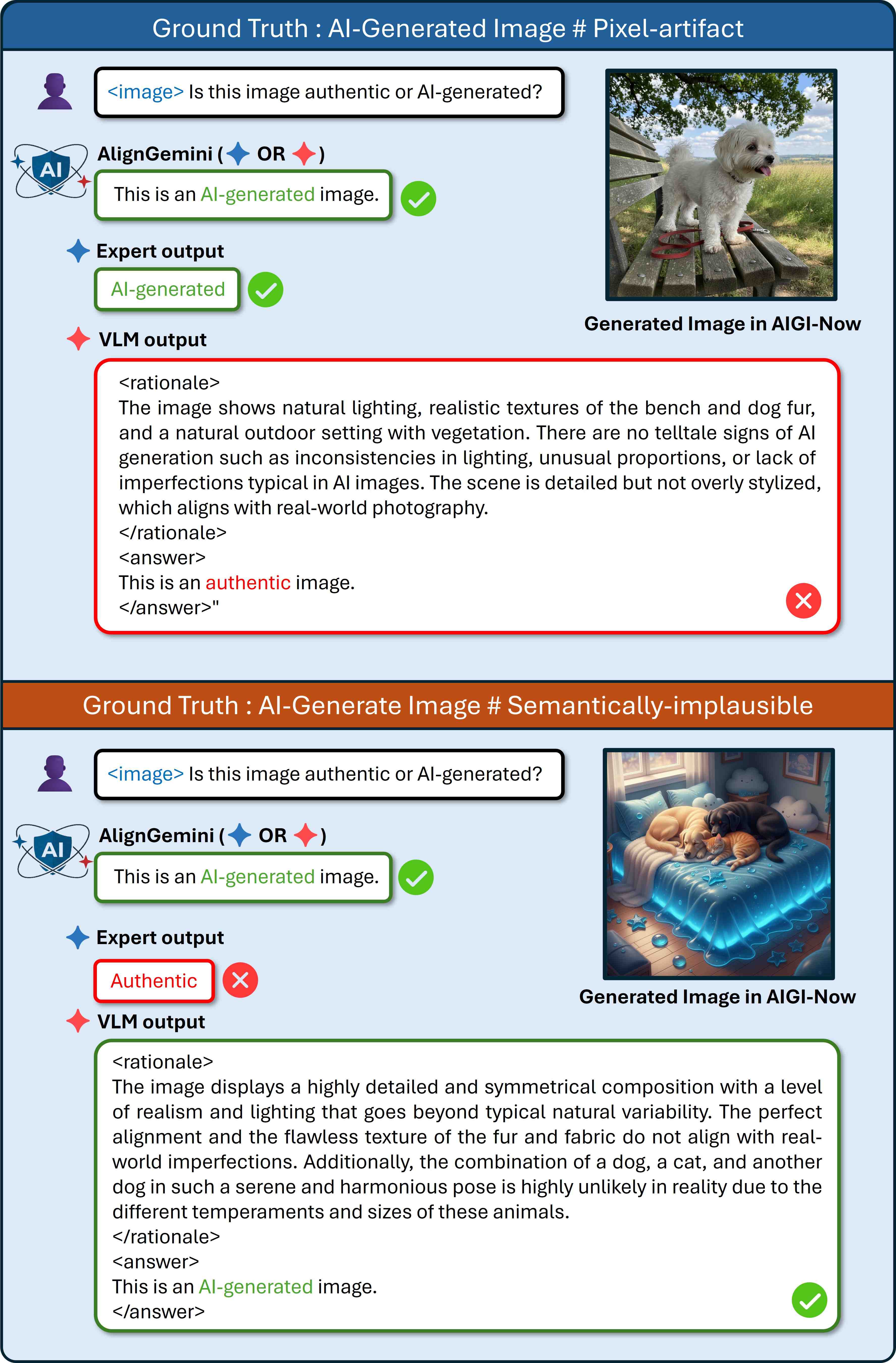}
\caption{Qualitative examples of {AlignGemini}'s predictions on AI-generated images from the AIGI-Now benchmark.}
\label{fig:visualization_chat_aigi_now}
\end{figure*}

\begin{figure*}
\centering
\includegraphics[width=0.8\linewidth]{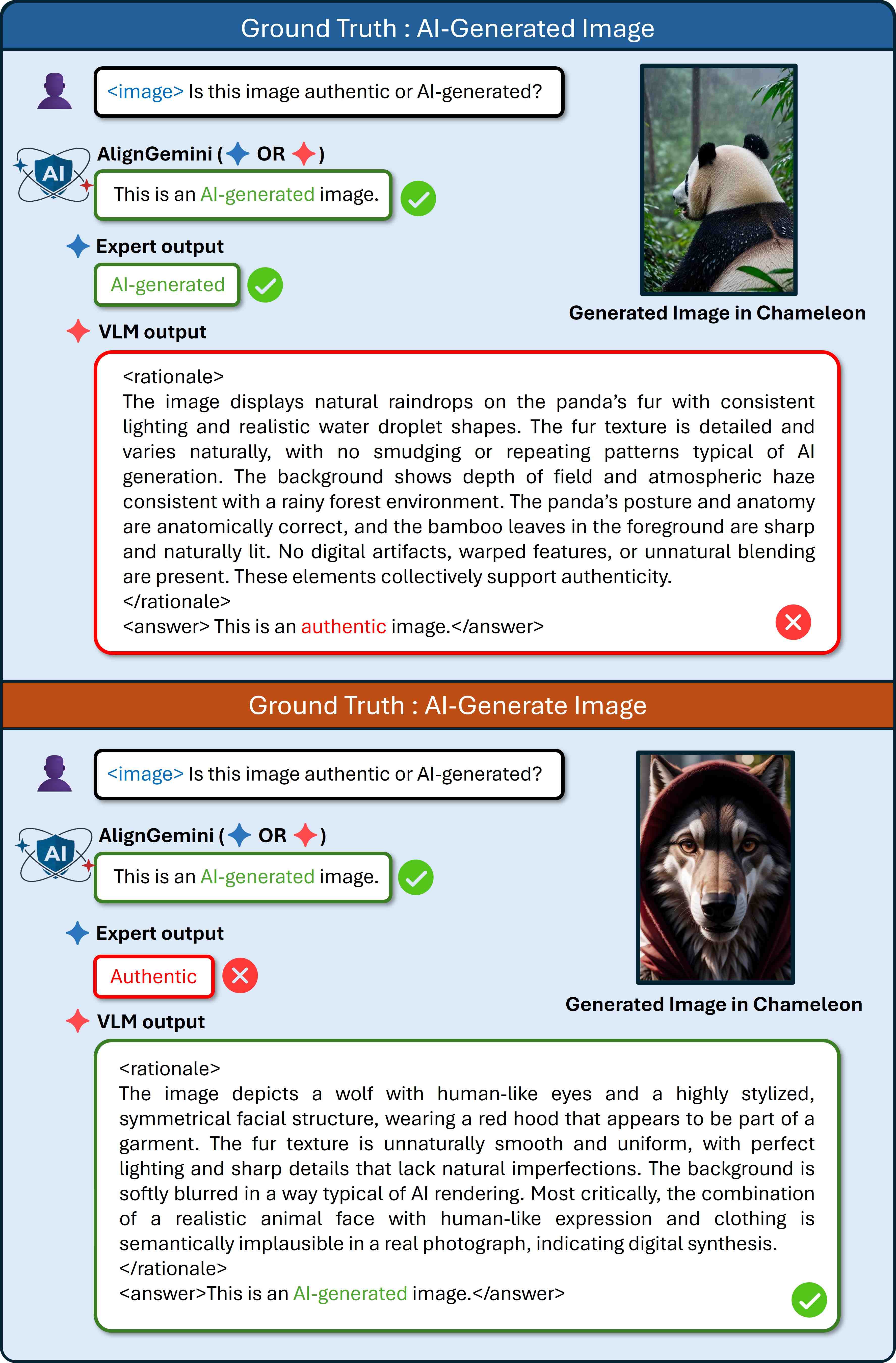}
\caption{Qualitative examples of {AlignGemini}'s predictions on images from the Chameleon benchmark.}
\label{fig:visualization_chat_chameleon}
\end{figure*}

\begin{figure*}
\centering
\includegraphics[width=0.8\linewidth]{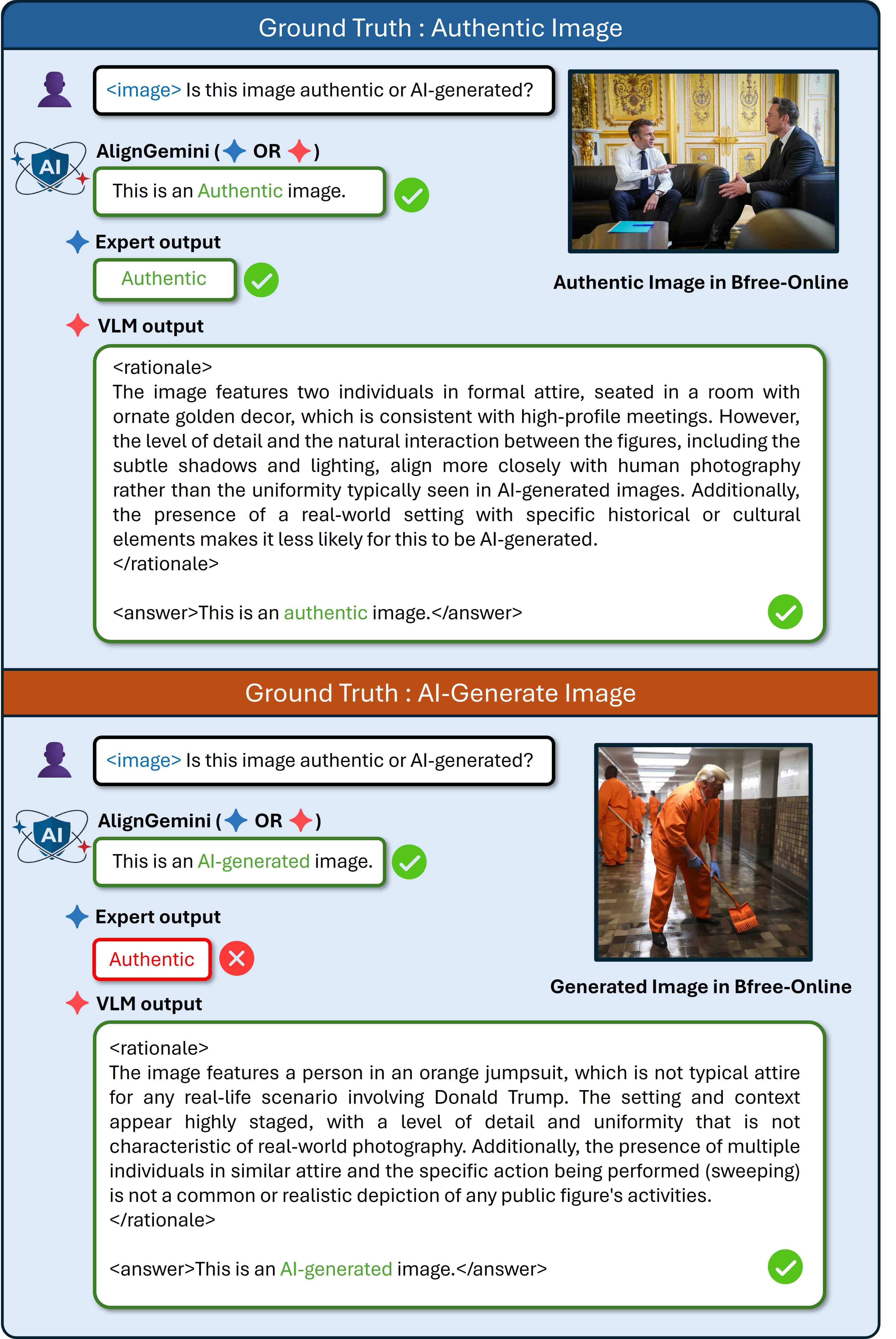}
\caption{Qualitative examples of {AlignGemini}'s predictions on images from the BFree-Online benchmark.}
\label{fig:visualization_chat_bfree}
\end{figure*}

\begin{figure*}
\centering
\includegraphics[width=0.8\linewidth]{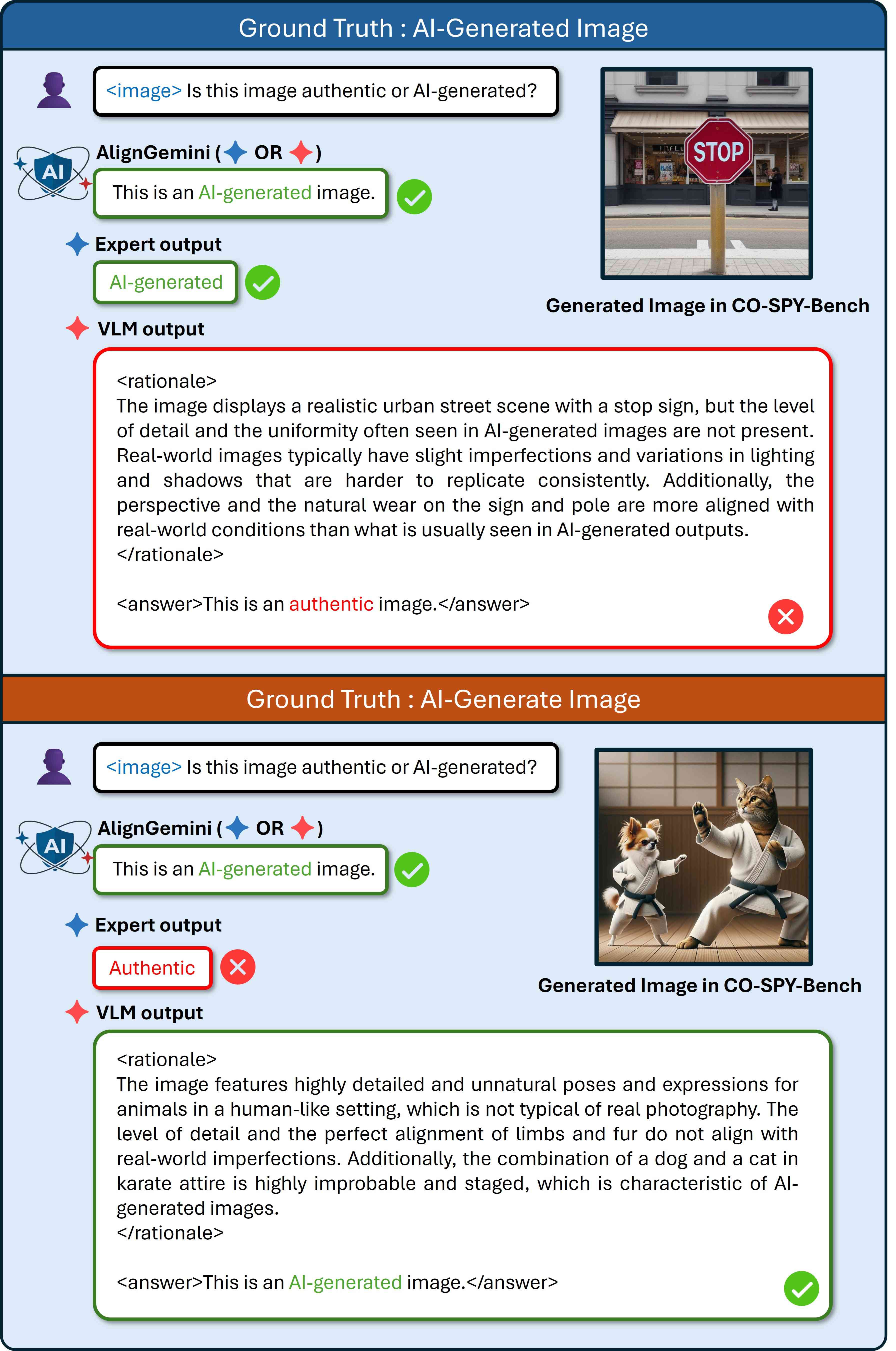}
\caption{Qualitative examples of {AlignGemini}'s predictions on images from the CO-SPY-Bench (in-the-wild) benchmark.}
\label{fig:visualization_chat_cospy}
\end{figure*}

\begin{figure*}
\centering
\includegraphics[width=0.8\linewidth]{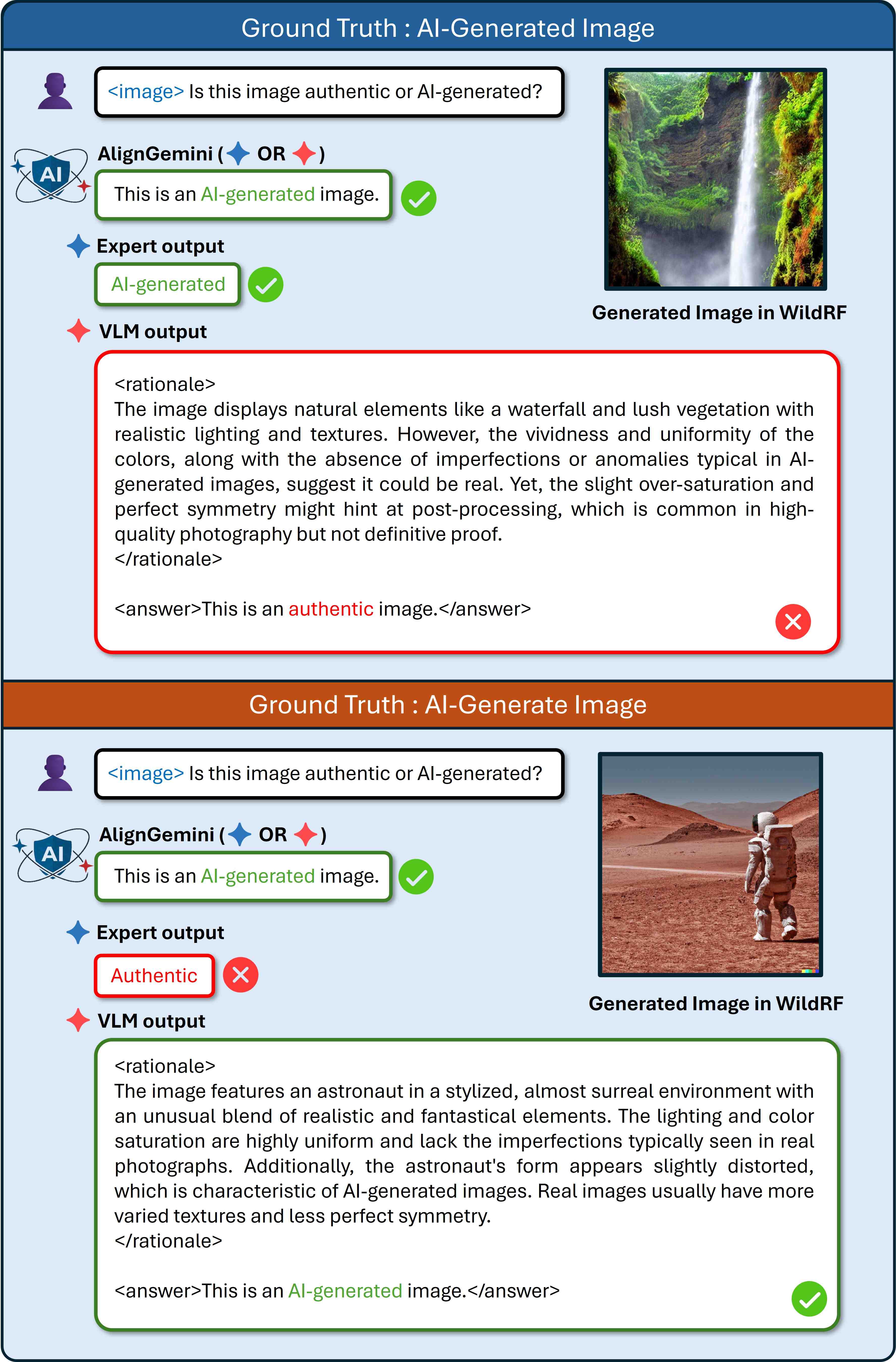}
\caption{Qualitative examples of {AlignGemini}'s predictions on images from the WildRF benchmark.}
\label{fig:visualization_chat_wildRF}
\end{figure*}

\begin{figure*}
\centering
\includegraphics[width=0.8\linewidth]{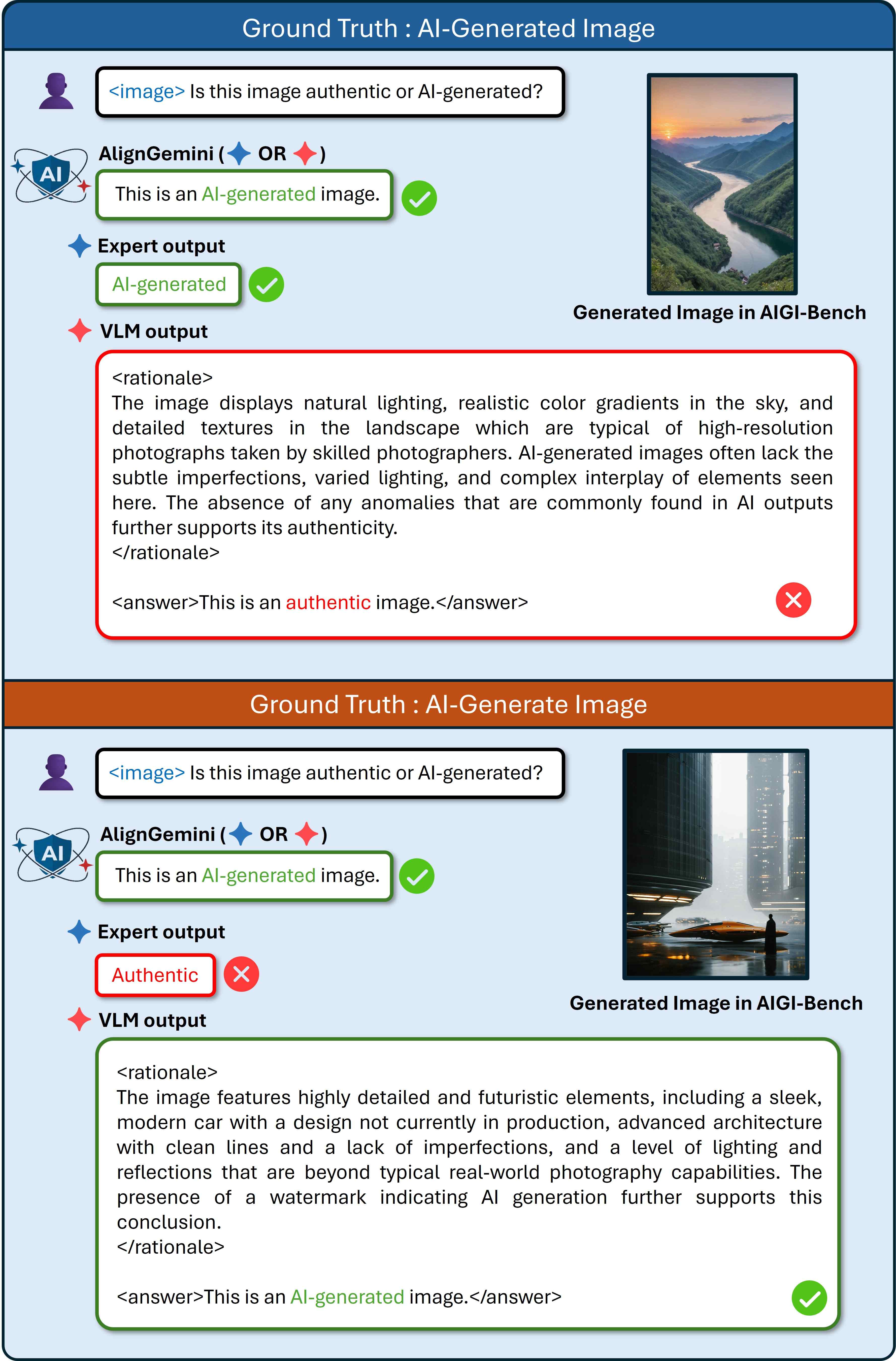}
\caption{Qualitative examples of {AlignGemini}'s predictions on images from the AIGI-Bench benchmark.}
\label{fig:visualization_chat_aigi_bench}
\end{figure*}

\paragraph{Visualization of AlignGemini's outputs.}
Figures~\ref{fig:visualization_chat_aigi_now}--\ref{fig:visualization_chat_aigi_bench} present qualitative examples of {AlignGemini} on various benchmarks.
{AlignGemini}'s final decision is obtained by applying an OR rule on the fake predictions of the expert branch and the VLM branch: an image is classified as real only if both branches predict it as real.
The visualizations show that the two branches provide complementary signals, with the expert model focusing more on pixel-level artifacts and the VLM emphasizing semantic inconsistencies, leading to more robust and generalizable discrimination.

\clearpage

\end{document}